\documentclass[sigconf]{acmart}

\usepackage{subfigure}

\pdfoutput=1

\def\BibTeX{{\rm B\kern-.05em{\sc i\kern-.025em b}\kern-.08emT\kern-.1667em\lower.7ex\hbox{E}\kern-.125emX}}
    
\copyrightyear{2019}
\acmYear{2019}
\setcopyright{acmlicensed}
\acmConference[CIKM '19] {The 28th ACM International Conference on Information and Knowledge Management}{November 3--7, 2019}{Beijing, China}
\acmBooktitle{The 28th ACM International Conference on Information and Knowledge Management (CIKM'19), November 3--7, 2019, Beijing, China}
\acmPrice{15.00}
\acmDOI{10.1145/3357384.3357870}
\acmISBN{978-1-4503-6976-3/19/11}

\begin{document}
\fancyhead{}

%
\title{DeepIST: Deep Image-based Spatio-Temporal Network for Travel Time Estimation}

\author{Tao-yang Fu}
\affiliation{%
  \institution{The Pennsylvania State University}
  \city{University Park} 
  \state{PA, 16802} 
  \postcode{16802}
  \country{USA}
}
\email{txf225@cse.psu.edu}

\author{Wang-Chien Lee}
\affiliation{%
  \institution{The Pennsylvania State University}
  \city{University Park} 
  \state{PA, 16802} 
  \postcode{16802}
  \country{USA}
}
\email{wlee@cse.psu.edu}

%

%
\begin{abstract}
Estimating the travel time for a given path is a fundamental problem in many urban transportation systems. However, prior works fail to well capture moving behaviors embedded in paths and thus do not estimate the travel time accurately. To fill in this gap, in this work, we propose a novel neural network framework, namely {\em Deep Image-based Spatio-Temporal network (DeepIST)}, for travel time estimation of a given path. The novelty of DeepIST lies in the following aspects: 1) we propose to plot a path as a sequence of ``generalized images'' which include sub-paths along with additional information, such as traffic conditions, road network and traffic signals, in order to harness the power of convolutional neural network model (CNN) on image processing; 2) we design a novel two-dimensional CNN, namely {\em PathCNN}, to extract spatial patterns for lines in images by regularization and adopting multiple pooling methods; and 3) we apply a one-dimensional CNN to capture temporal patterns among the spatial patterns along the paths for the estimation. Empirical results show that DeepIST soundly outperforms the state-of-the-art travel time estimation models by 24.37\% to 25.64\% of mean absolute error (MAE) in multiple large-scale real-world datasets.\end{abstract}

%
%


%

%

%
\maketitle

\section{Introduction}

With the proliferation of GPS-enabled devices and the needs of location-aware applications, enormous amounts of trajectory data are being generated at an unprecedented speed. The massive trajectory data give opportunities to carry out various mining tasks, such as trajectory classification, trajectory clustering, {\em travel time estimation}, trajectory outlier detection, etc, in support of urban planning and transportation system operations and management. Among these mining tasks, {\em estimating the travel time for a given path}, which can be denoted by a sequence of connected road segments in a road network, is fundamental to many urban transportation systems, e.g., route planning, freight management, navigation, traffic monitoring and ride sharing~\cite{zhang2018deeptravel,wang2018will}. It is a nontrivial problem because 1) the travel time is impacted by many factors along the path, such as the dynamics of traffic condition, the types of road segments and intersections, and the traffic signals (e.g., traffic lights, stop signs and crossings); and 2) the spatio-temporal moving behaviors of {\em mobile road users} (also refer to as {\em mobile users}), while being captured in trajectories, are not well understood and used.

\begin{figure}[t]
	\centering
	\includegraphics[scale=0.31]{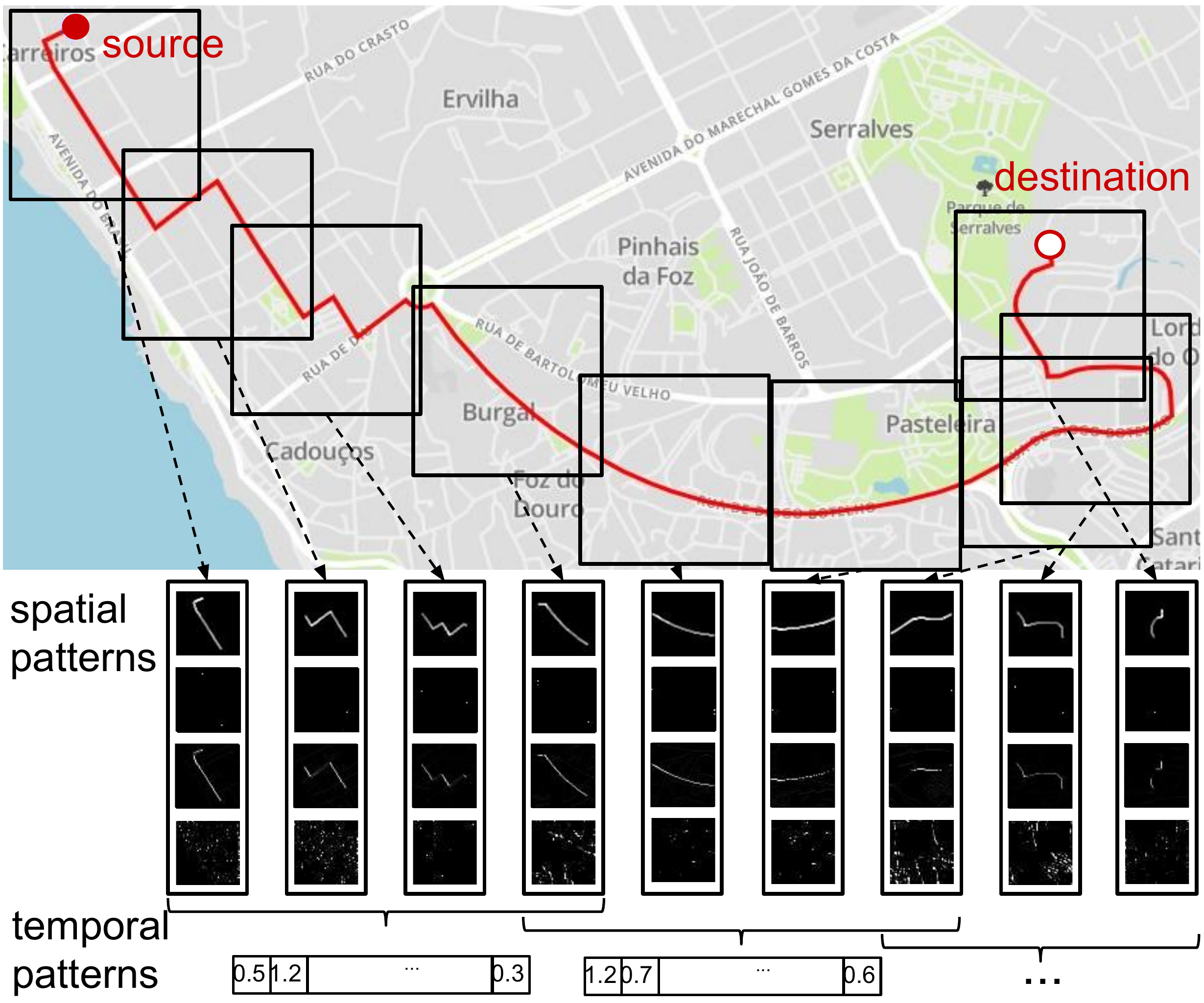}
	\vspace{-2mm}
	\caption{Spatio-temporal moving behaviors in a path}
	\label{fig:example}
	\vspace{-7mm}
\end{figure}

Let's use the example trajectory in Figure~\ref{fig:example} to illustrate the effect of the various factors and the moving behaviors of a vehicle driver who prefers to drive on main streets and highways to arrive at the destination fast. The (red) path shows that i) the driver first wanders on local streets from the source to approach the closest intersection to a main street, during which she encounters multiple intersections, stop signs and traffic lights, takes turns frequently, and thus moves slowly; ii) then she moves fast on that main street towards the destination and seldom stops; and finally iii) she wanders again on local streets to the destination. While we can roughly recognize the general moving behaviors of the driver, manually identifying various potential moving behaviors of drivers for travel time estimation is impractical as the moving behaviors in the real-world are really complex. Therefore, research on developing effective techniques to exploit the abundant collected trajectory data available nowadays is a mandate. Ideally, an effective travel time estimation technique would {\em automatically} learn important moving patterns (in both of the spatial and temporal dimensions) along paths corresponding to various major factors, leading to accurate estimation of travel time for paths. For example, as illustrated at the bottom of Figure~\ref{fig:example}, (visible) spatial moving patterns and temporal patterns expressed in a vector may be captured. The four spatial patterns signal the ``moving path'' and ``source and destination'' of the path, the traffic conditions along the path and the intersections of the underlying road network. Moreover, the temporal patterns denote some implicit factors, e.g., ordering relationship among a sequence of spatial patterns.


The problem of travel time estimation has been widely studied in the past~\cite{rice2004simple,lv2015traffic,ma2015long, yang2018pace}.
However, they are usually designed in an ad hoc fashion, based on some strong assumptions, which results in poor accuracy. Recently, deep learning techniques have been explored for travel time estimation~\cite{zhang2018deeptravel,wang2018will,wang2018learning}. DeepTravel~\cite{zhang2018deeptravel} represents a trajectory by a sequence of geographical grid cells, where manually-craft features are extracted to serve as inputs for Long Short Term Memory (LSTM) neural network to train a model for travel time estimation. WDR~\cite{wang2018learning} manually designs a number of features for query paths and proposes an ensemble regression model for travel time estimation. DeepTTE~\cite{wang2018will} employs a sliding window to transform a path as a sequence of windows of sample points and extract features from them by a shallow 
network to train an LSTM model for travel time estimation. Through our analysis, we observe two major pitfalls in these existing deep learning techniques: 1) the spatial features extracted from trajectories do not effectively reflect the complex spatial moving behaviors (patterns) of mobile users due to manually-craft features or shallow networks; and 2) the temporal patterns among the spatial moving patterns are not well captured due to the vanishing and exploding gradient issues inherited from the use of LSTM~\cite{pascanu2013difficulty, sutskever2014sequence, greff2017lstm}.

\begin{figure}[ptb]
	\centering
	\includegraphics[scale=0.38]{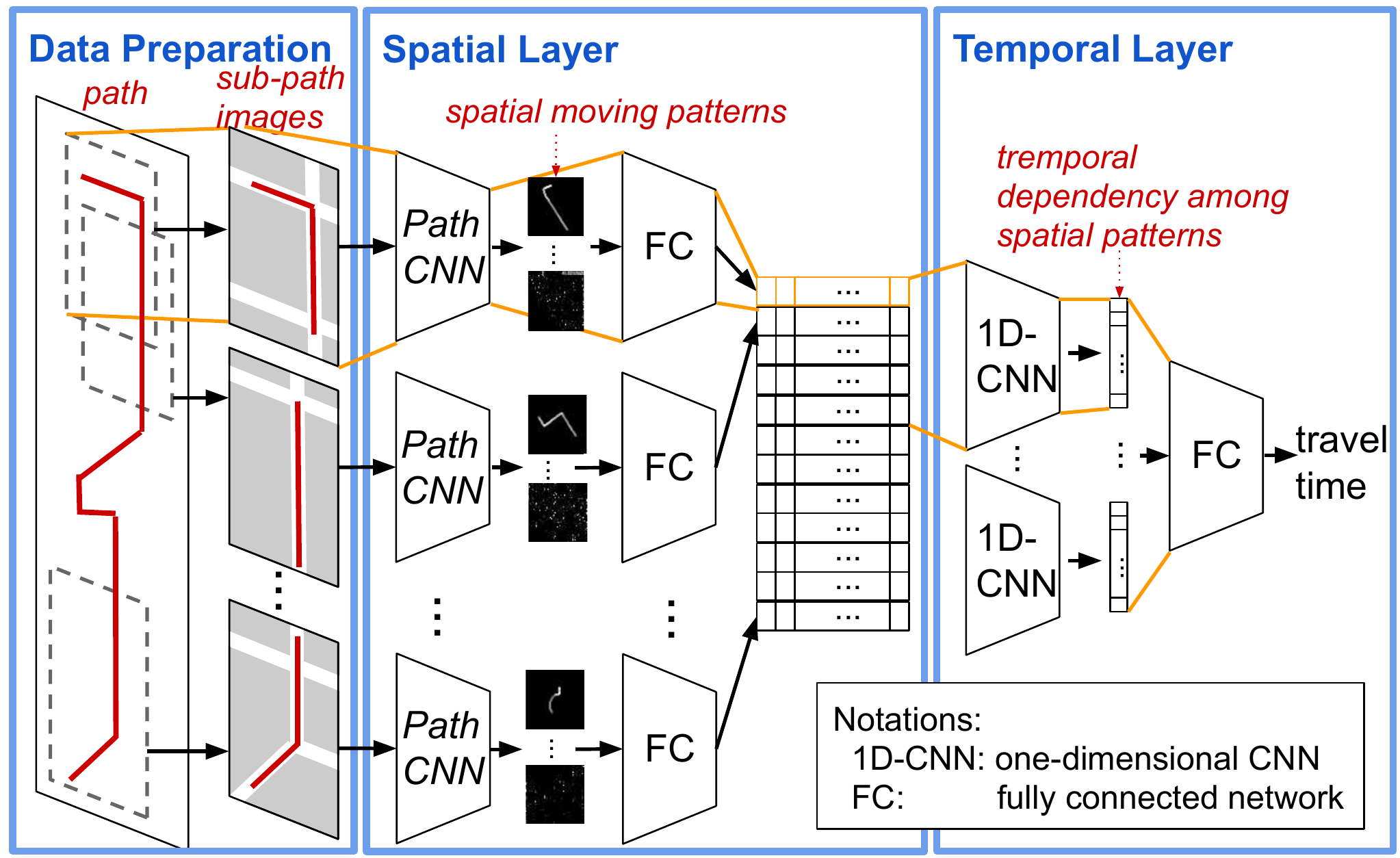}
	\vspace{-2mm}
	\caption{The architecture of DeepIST}
	\label{fig:sys_flow}
	\vspace{-6mm}
\end{figure}

To address the above-mentioned issues, in this paper, we propose to represent the movements of mobile users on road networks (i.e., paths) as {\em generalized images} in order to harness the proved power of \textit{convolutional neural network model (CNN)} to capture the complex moving patterns along paths for travel time estimation. CNN models, a renown class of neural network models for image processing, are designed to automatically learn spatial hierarchies of features in different levels of details from image data. To realize this new idea, an approach is to plot a path into one image, where useful information corresponding to various factors (e.g., estimated traffic condition along the path and the underlying road network) are plotted in individual channels of the image, and then apply CNN to extract features from the generated image for travel time estimation. Note that the generated "image" is actually represented as a three-dimensional tensor where the number of channels is not limited by three (thus the image is generalized). However, to process large-size images with high resolution, this approach is impractical due to limited computing resources, e.g., memory of GPUs, in an average server. Moreover, while CNN is powerful for modeling spatial patterns in images, it's good mainly at capturing features for image textures but not for lines (i.e., paths in our case)~\cite{geirhos2018imagenet}. Further, it is not known for modeling temporal properties.

To address the above challenges, we propose a new framework, namely {\em Deep Image-based Spatio-Temporal network (DeepIST)}, for travel time estimation of given paths. As shown in Figure~\ref{fig:sys_flow}, DeepIST is a three-layer framework consists of the following components: (1) {\em Data preparation layer}: to address the resolution issue of images and maintain the temporal properties of paths, it employs a sliding window (with a moving distance $w$ as the window size) over a path from the source to the destination to plot the sub-path in each window into an image, which also contains the additional information; (2) {\em Spatial layer}: given a sequence of images generated from a path, we propose a new two-dimensional CNN model, called {\em PathCNN}, which regulates classical convolutions for extracting spatial features of lines, applies multiple pooling methods adapting for different types of information in images, 
extracts spatial moving patterns embedded in the sub-paths, 
and consequently transforms the path into a sequence of spatial moving patterns; and (3) {\em Temporal layer}: given a sequence of spatial moving patterns, it applies a one-dimensional CNN (1D-CNN) model~\cite{kim2014convolutional} to capture local temporal patterns among consecutive spatial patterns along the path for travel time estimation. To the best of our knowledge, this is the first attempt to treat paths as sequences of images and seamlessly capture both spatial moving patterns and their temporal patterns in paths for travel time estimation.

The major contributions of this work are summarized as follows.

\begin{itemize}
\vspace*{-0.1cm}
\item {\bf Novel ideas for travel time estimation.}
This work analyzes the challenges of travel time estimation for paths and proposes to treat paths as images in order to exploit CNN models to overcome those challenges.


\item {\bf A new framework for travel time estimation}. We propose DeepIST to estimate the travel time for paths. DeepIST consists of a data preparation layer to generate sequences of sub-path images from query paths, and two layers of CNN-based models to capture spatial moving patterns and temporal patterns among those spatial patterns, respectively. 

\item {\bf A new 
CNN for spatial patterns of lines}. We propose PathCNN which adds penalties to enforce convolutions to extract spatial features of lines (in sub-paths and road networks).
Moreover, it uses multiple pooling methods adapting for different types of information in the generated images.

\item {\bf Empirical evaluation using real-world data}. We evaluate DeepIST by conducting a comprehensive evaluation using two large-scale real-world trajectory datasets in comparison with state-of-the-art techniques, including road-based, path-based and deep learning methods. Empirical result shows DeepIST soundly outperforms all existing models.

\end{itemize}

The rest of this paper is structured as follows. We review the related work in Section 2 and provide problem definition and analysis in Section 3. We detail the proposed DeepIST framework in Section 4 and show experiment results in Section 5. Finally, we conclude the paper in Section 6 and discuss future research directions.
\section{Related Work}

Existing works on travel time estimation generally fall into three categories: road-based, path-based and learning-based techniques.

\noindent{\bf Road-Based Travel Time Estimation. }Works in this category estimate the travel time (or speed) on individual road segments in a road network. In turn, the travel time on a query path is estimated by summing up the travel times of road segments in the path. 
Techniques fallen in this category have explored various mechanisms and learning models, including loop detectors~\cite{jia2001pems, rice2004simple}, probabilistic distribution models~\cite{de2008traffic}, dynamic Bayesian network~\cite{hofleitner2012learning}, spatial-temporal Hidden Markov Model~\cite{yang2013travel}, support vector regression~\cite{asif2014spatiotemporal}, ensemble models~\cite{wang2016etcps}, stacked auto-encoder~\cite{lv2015traffic} and LSTM~\cite{ma2015long}. As these prior studies do not consider the interactions and correlations among road segments, 
they miss high-level moving patterns among road segments and thus have difficulty achieving high accuracy,
especially when local errors on individual road segments are accumulative~\cite{jenelius2013travel}. Some works do try to model interactions and correlations between the adjacent road segments~\cite{yang2013travel, wang2016etcps}. However, as they still focus on individual road segments or pairs of adjacent road segments, these works face the same pitfalls mentioned above. 


\noindent{\bf Path-based Travel Time Estimations. }To address the above-mentioned issues, path-based approaches use common characteristics of paths (or their sub-paths) to find similar paths from historical data for travel time estimation. Some works mine frequently passed sub-paths from historical path data~\cite{luo2013finding} or extract common sub-paths between the query path and historical paths~\cite{rahmani2013route} or select an optimal set of sub-paths that together cover the query path~\cite{yang2018pace} to make estimation by averaging the travel time on the whole query path. However, these approaches do not lead to good result when the extracted sub-paths do not match well with that of the query path. Based on an assumption that paths with close sources and destinations share the same/similar route and thus have similar travel time, some works find trajectories with nearby source and destination or nearby trajectories to the query path to derive the travel time distribution for estimation~\cite{wang2016simple, li2018multi}. However, these approaches do not achieve good accuracy when their assumption fails. In~\cite{yuan2013t}, Yuan et al. build a landmark graph from trajectory data, based on top-k frequently traversed road segments, where the travel times between landmarks are derived. A query path is transformed into a sequence of landmarks to estimate its travel time by summation of landmark-to-landmark travel times. However, query paths not well covered by the landmark graph are not estimated accurately. In~\cite{wang2014travel}, Wang et al. apply tensor decomposition to derive travel time of unseen sub-paths in historical trajectory data. Thus, the travel time of a query path passing by these sub-paths can be estimated. However, it still suffers from the data sparsity issue appearing in rarely traveled sub-paths.

\noindent{\bf Learning-based Techniques. } Deep learning techniques have recently been proposed for travel time estimation~\cite{zhang2018deeptravel, wang2018learning, wang2018will}. {\em DeepTravel}~\cite{zhang2018deeptravel} represents a query path as a sequence of geographical grid cells, where a number of manually-craft features are extracted, including geographic location, timestamp, driving state, average speed, and the number of trajectories in a cell in recent minutes and days. Using the sequence of extracted features as input to an LSTM model, DeepTravel aims to capture the temporal patterns in a path to estimate the travel time. However, performing feature engineering manually is not only labor-intensive but also difficult to capture numerous complex moving patterns. Moreover, it's tricky to select a proper cell size. Too large a cell size aggregates many sample points, resulting in loss of details in moving behaviors; too small a size leads the learning process to suffer from data sparsity. {\em Wide-Deep-Recurrent (WDR)}~\cite{wang2018learning} designs a number of manually-craft features for a query path, including spatial, temporal, traffic, personalized and augmented information and propose an ensemble regression model, consisting of a wide, a deep network and an LSTM model, to estimate the travel time of the path. However, also adopting manually designed features, it faces the same pitfalls mentioned above. {\em DeepTTE}~\cite{wang2018will} employs a sliding window to transform a path as a sequence of windows, each of which contains consecutive sample points along the query path. By extracting $k$ spatial features from the sample points in each window (represented as geographic vectors by non-linearly mapped from their geographic locations), DeepTTE applies the LSTM model to capture the temporal patterns in 
those spatial features to estimate the travel time. While DeepTTE employs fully connected networks for feature extraction, it does not effectively capture spatial features of a window~\cite{krizhevsky2012imagenet}, not to mention the complex moving patterns of mobile users.

As mentioned, the approaches adopted by DeepTravel, WDR and DeepTTE may not effectively capture the complex moving patterns in paths. In our work, we demonstrate that DeepIST, by harnessing the strengths of CNN models, is able to {\em automatically} (without human intervene) capture the complex spatial moving patterns and their temporal patterns along paths for travel time estimation.
\section{Research Problem and Challenges}

In this section, we introduce important terms, describe the targeted research problem and discuss the challenges.

\noindent {\bf Road Network}. A road network is a directed graph $G = (V,E,\Phi,\Psi)$, where $V$ is a set of nodes denoting intersections and each node $v \in V$ contains a geographic location $(v.lng, v.lat)$ (i.e., longitude and latitude); $E \subseteq V \times V$ is a set of directed edges denoting road segments and each edge $e \in E$ corresponds to a road segment from a start node $e.s = v \in V$ to an end node $e.e = v'\in V$, where $v' \neq v$.; and $\Phi : V \to A$ and $\Psi : E \to R$ are type mapping functions for nodes and edges, respectively, e.g., an intersection has a traffic signal and a road segment is a highway.

\noindent {\bf Path}. A path $T$
is a sequence of road segments in a road network, i.e., $T= \{e_1, ..., e_{|T|}\}$
where $e_1$ and $e_{|T|}$ may be partial road segments.

\noindent {\bf Trajectory}. A trajectory is a sequence of spatio-temporal sample points generated from the movement of a mobile user on a path, where a sample point contains a location (i.e., longitude and latitude) and a timestamp.

Owing to the advent in positioning technology, trajectory data of mobile users can be easily collected and used for transportation analytics, including travel time estimation. Historical trajectories, which can be mapped to paths on road networks using map matching techniques, provide ground truth for learning a predictive model for 
travel time estimation of paths. In the following, we formally state our research goal.

\noindent {\bf Travel Time Estimation for Paths }. Given a path dataset (converted from historical trajectories) $D=\big\{(T_i, s_i, a_i)\big\}_{i=1}^{|D|}$, where $T_i$ is the $i$-th path, $s_i$ is the departure time and $a_i$ is the arrival time, we learn a neural network based model to estimate the travel time on a path 
by extracting spatial moving patterns and their temporal patterns from $D$ 
for training. During the test phase, given a query path $T_q$ and a departure time $s_q$, which is not observed in $D$, 
the travel time of $T_q$ is estimated.

Note that the path data, inherited from trajectories, may also contain arrival time information at trajectory sample points,
which are not available during test phase. 
To avoid the learned model simply counting the number of sample points to exploit the fixed time gap between consecutive sample points for travel time estimation, we do not directly use the arrival time at trajectory sample points in training. 
In this work, we propose a framework, DeepIST, to tackle the problem of travel time estimation for paths
by treating a path as a sequence of images and exploiting the power of CNN models to capture the spatial moving patterns and their temporal patterns in paths for the estimation. To implement the DeepIST framework, we face two new challenges: (1) {\em Data preprocessing.} To realize our idea, a proper way to transform a path as a sequence of sub-paths plotted in images is essential for effective and efficient model learning. How to properly split a path into a sequence of sub-paths which contain sufficient local spatial information? How to plot a sub-path in an image, with useful information, such as the estimated traffic condition, the traffic signals and the underlying road network? (2) {\em Spatial and temporal pattern mining.} A well designed model is critical for effective and efficient learning to capture the spatial and temporal patterns in paths for the intended estimation. We propose to design an end-to-end CNN-based model consisting of two layers: (i) the spatial layer, where we propose a two-dimensional CNN based model, named \textit{PathCNN}, aiming to extract the spatial moving patterns of the sub-path in each image, and (ii) the temporal layer, which exploits 1D-CNN to capture the local temporal patterns among consecutive sub-paths. How to design a model consisting of the two layers? How to regulate the convolutions to capture spatial patterns of sub-paths? What is the proper loss function for model learning? These are research questions arising in the design of DeepIST.
\section{The DeepIST Framework}

As introduced earlier in Figure~\ref{fig:sys_flow}, the DeepIST framework consists of three layers: {\em Data preparation layer}, {\em Spatial layer} and {\em Temporal layer}. In the section, we detail our design in each layer.

\subsection{Data Preparation Layer}

\begin{figure}[t]
	\centering
	\begin{minipage}[t]{0.56\linewidth}
		\centering
		\includegraphics[width=1\linewidth]{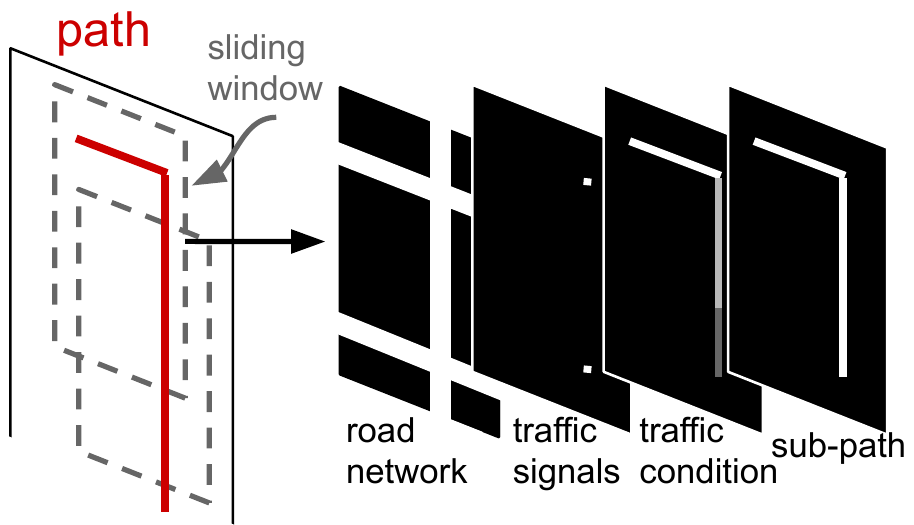}
		\caption{Data preparation layer}
		\label{fig:data}
	\end{minipage}
	\hspace{25pt}
	\begin{minipage}[t]{0.31\linewidth}
		\centering
		\includegraphics[width=1\linewidth]{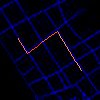}
		\caption{An image}
		\label{fig:data_example}
	\end{minipage}
	\vspace{-3mm}
\end{figure}

As mentioned earlier, simply plotting the whole path into one image for CNN models to estimate the travel time is not a good idea due to the 
excessive computing resources required for processing large-size images with high resolution. Moreover, large-size images usually require more layers in deep learning models to process, thus incurring expensive training time. On the other hand, small-size images with low resolution may not provide sufficient details of the path to achieve a good performance. Therefore, for a given path, instead of generating one image for the whole path, we propose to employ a sliding window over the path to generate a sequence of windows (e.g., the black squares shown in Figure~\ref{fig:example}), each of which contains a sub-path. Accordingly, we plot the sub-path in each window to generate an image. This not only addresses the resolution issue but also makes the estimation focusing on nearby area of the path, which is more discriminative for the estimation, than distant area of the path.
More specifically, we employ a sliding window over the path, with the window size $w$ and sliding step $s$, to generate a sequence of windows, each of which contains a $w$ kilometer sub-path, i.e., $\{\tau_1, ..., \tau_n\}$. As a result, a path is represented by a sequence of images containing ordered sub-paths of the path $I=\{I_1, ..., I_n\}$. The sliding step $s$, ($w \ge s > 0$), configures the length of overlaps between adjacent windows. With maximal sliding step $s=w$, non-overlapped windows are generated, which may cause some moving patterns between adjacent windows to be missed. On the other hand, a small step generates a long sequence of windows for the given path (due to overlaps between adjacent windows) which leads to expensive training time. To generate an image, we take a geographic area $r_{lng} \times r_{lat}$, where $r_{lng}$ and $r_{lat}$ are the geographic range in longitude and latitude, surrounding the sub-path. Next we locate the sub-path $\tau_i$ under processing at the center of the area, which is calculated by $\left( \frac{e_1.s.lng+\sum_{e_j \in \tau_i}{e_j.e.lng}}{|\tau_i|}, \frac{e_1.e.lat+\sum_{e_j \in \tau_i}{e_j.e.lat}}{|\tau_i|}\right)$, to generate an image of fixed resolution, i.e., with $k \times k$ pixels.

In addition, we propose to not only plot a sub-path of the path into an image, but also additional factors which may affect the travel time of a path: 1) the traffic condition on the road segments along the sub-path; 2) the underlying road network in the geographic area of the image; and 3) the traffic signals (e.g., stop signs, traffic lights and crossing) nearby the sub-path. There are other potential factors which may also affect the travel time, such as the terrain in the area, the weather condition, the speed limitation of road segments and the personal driver preference, but here we focus on the above-mentioned three factors and leave others for future work. More specifically, in addition to the sub-path itself, information corresponding to each factor is plotted in an individual channel to make a ``generalized image''.\footnote{The number of potential channels is not limited, so our approach can be generally extended to incorporate more factors, which is actually represented as a three-dimensional tensor.}
While plotting the sub-path provides moving distance and shape of movement, we argue that {\em moving speed} may serve as an indicator of traffic condition for travel time estimation. In this work, 
we use a state-of-the-art~\cite{ma2015long} method that estimates the hourly traffic condition of road segments. The value of traffic condition on a road segment is normalized to 0 to 1 by the maximum speed of the whole dataset to plot the road segment on the image. On the other hand, we argue that plotting the underlying road network may provide useful information about intersections encountered and types of road segments along the sub-path. It takes a longer time to travel if more intersections are encountered along a sub-path. In the plot, 
we differentiate the types of road segments by the width of lines plotted, e.g., two pixels for highways and one pixels for others, where the types of road segments are obtainable from digital maps such as OpenStreetMap~\cite{openstreetmap}. Finally, plotting traffic signals (e.g., stop sign, traffic light or crossing) as pixels may help to indicate where needs to take longer time along a sub-path.

In summary, as shown in Figure~\ref{fig:data}, for a given path $T$, in the data preparation layer, we employ a sliding window to split $T$ as a sequence of windows containing ordered sub-paths $\{\tau_1, ..., \tau_n\}$. Then, for each window, we generate an image by plotting a sub-path along with additional factors plotted in individual channels. Only for visualization, an image shown in 
Figure~\ref{fig:data_example} plots 1) the sub-path in Red, 2) the estimated traffic condition of road segments along the sub-path in Green, and 3) the underlying road network in the same geographic area of the image in Blue. The traffic signals are not displayed due to the 3-channel format of a normal image. 

\subsection{Spatial Layer}

After generating a sequence of images for a given path in the previous layer, the spatial layer extracts spatial moving patterns of the sub-path in each image. Inspired by~\cite{krizhevsky2012imagenet}, we design a two-dimensional CNN model, called PathCNN, which 
adopts multiple pooling methods for adapting the heterogeneous types of information i.e., sub-path, traffic condition, underlying road network, and traffic signals, in images. We argue that {\em max} pooling is suitable for binary values in sub-path, underlying road network, and traffic signals, while {\em average} pooling is suitable for numeric values in  traffic condition. Therefore, we apply both pooling schemes in our model. Moreover, as classical CNN models mainly capture features for image textures instead of lines (e.g., sub-paths in our case)~\cite{geirhos2018imagenet},
PathCNN regulates its convolutions to better capture spatial features of sub-paths. 

\begin{figure}[t]
	\centering
	\includegraphics[scale=0.47]{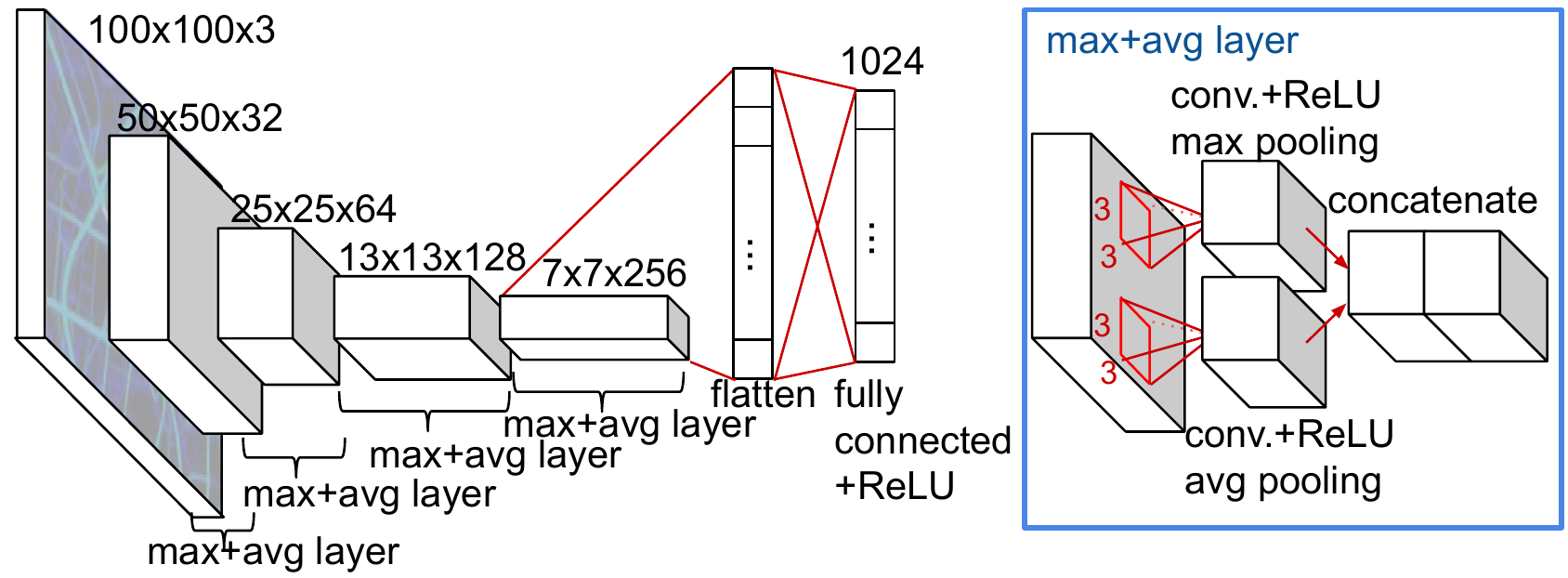}
	\caption{Spatial Layer: The PathCNN Model}
	\vspace{-2mm}
	\label{fig:spatial}
	\vspace{-2mm}
\end{figure}

\noindent{\bf Model Architecture. }As shown in Figure~\ref{fig:spatial}, the proposed PathCNN model contains $M$ layers (e.g., $M$=4), named max+avg layers, followed by a flatten layer and a fully-connected layer. More specifically, for each image $I_i$ in the image sequence $I = \{I_1, ..., I_n\}$ of a path $T$, a three-dimensional tensor $I_i \in R^{k \times k \times d}$ where $k\times k$ is the size of the image and $d$ is the number of channels used in the image (e.g., $k$=100 and $d$=4). PathCNN takes $I_i$ as input $x_i^0$ and feeds $x_i^0$ into $M$ max+avg layers. In each max+avg layer $m$, PathCNN has two separate convolutional layers, which are followed in parallel by a max pooling layer and an average pooling layer, respectively. Then, the outputs of the two pooling layers are concatenated as the final output of the parallel pooling layer $m$. In each convolutional layer, it uses padding for location at boundaries of the image to maintain the dimension of representations. The transformation at each layer $m$ is derived as follows:

\vspace{-4mm}
\begin{center}
\begin{align*}
    x_{i_{max}}^m &= max\_pooling\big(f(x_i^{m-1} \ast W^m_{2D_{max}} + b^m_{2D_{max}})\big) \\
    x_{i_{avg}}^m &= avg\_pooling\big(f(x_i^{m-1} \ast W^m_{2D_{avg}} + b^m_{2D_{avg}})\big) \\
    x_i^m &= (x_{i_{max}}^m, x_{i_{avg}}^m)
\end{align*}
\end{center}

\noindent where $\ast$ denotes the convolutional operation and $f(\cdot)$ is an activation function. Here, we use the ReLU function for activation, i.e., $f(z) = ReLU(z) = max(0, z)$. $W^m_{2D_{max / avg}}$ and $b^m_{2D_{max / avg}}$ are the weights and biases of convolutions in the $m^{th}$ convolution layer for max pooling and for average pooling, respectively. We then use $max\_pooling(\cdot)$ and $avg\_pooling(\cdot)$ to reduce the dimension of the representations after applying convolutions. The number of convolutioins in each convolutional layer in each max+avg layer is denoted as $c_{2D}^m$ and the size of each convolution is denoted as $f_{2D}$, (e.g., $c_{2D}^m$=16, 32, 64, 128 for $m$=1,2,3,4, respectively and $f_{2D}$=3, as shown in Figure~\ref{fig:spatial}).

There are other potential ways to design the max+avg layer. For example, instead of two convolutional layers, using one followed by both of the max pooling and average pooling layers. We evaluate the different designs in the experiments.

After $M$ max+avg layers, we use a flatten layer to transform the output $x_i^M$ to a feature vector $\overline{s_i}$, where the length of $\overline{s_i}$ depends on the size of input image. Finally, we use a fully-connected layer to apply non-linear transformation to reduce the dimension of $\overline{s_i}$ to a feature vector $s_i \in R^{\lambda}$ (e.g., $\lambda$ = 1024 in Figure~\ref{fig:spatial}), which represents the spatial moving patterns in $\tau_i$ derived as $s_i = f\big(W \overline{s_i} + b\big)$, where $W$ and $b$ are the weights and biases of the fully-connected layer. Note that we adopt dropout mechanism~\cite{krizhevsky2012imagenet} in the fully-connected layer to avoid overfitting and make the model learning more robust.

Finally, for a given image sequence $I = \{I_1, ..., I_n\}$ of a trajectory $T$, after each image $I_i$ is processed by PathCNN into a feature vector $s_i$, we concatenate each $s_i$ as a matrix $S \in R^{n \times \lambda}$, which represents a sequence of spatial moving patterns along $T$.

\begin{figure}[t]
	\centering
	\includegraphics[scale=0.75]{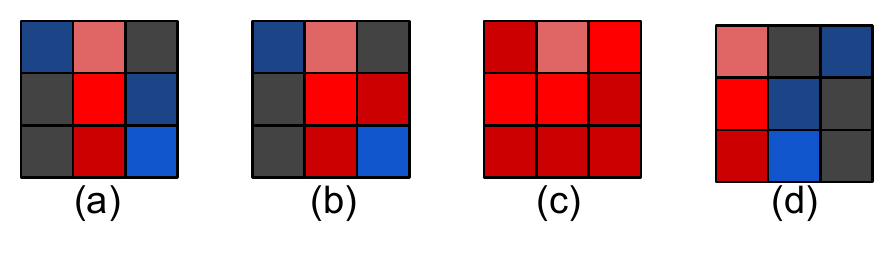}
 	\vspace{-3mm}
	\caption{Regularizations for Paths}
	\label{fig:regularize}
	\vspace{-5mm}
\end{figure}

\noindent{\bf Regularization for Paths. }As mentioned earlier, although CNN is powerful for modeling spatial patterns in general images, it captures features for textures well but not good at lines~\cite{geirhos2018imagenet}. Although 
the regularization methods for convolutions in CNN, e.g., $L_1$ and $L_2$ regularization, dropout~\cite{hinton2012improving}, batch normalization~\cite{ioffe2015batch}, and weighted dropout~\cite{hou2019weighted}, have been well studied, they focus on improving the resistance to the overfitting in convolutional layers rather than enforcing the convolutions to capture features for lines.

Conceptually, our idea is that a convolution well capturing features of lines would satisfy two criteria: 1) a convolution mainly captures lines; and 2) lines captured should be located at the center of the convolution. For examples, Figure~\ref{fig:regularize} shows four $3 \times 3$ convolutions illustrated as heat maps (where red denotes positive value, blue denotes negative value and black denotes zero). Convolution (a) and (b) are examples of ideal convolutions which capture one straight line and two crossed lines (i.e., a T-junction), respectively. On the other hand, convolution (c) does not capture lines. Finally, although convolution (d) captures one straight line, the line captured is not located at the center of it.

While learning to regulate convolutions in each convolutional layer to approach the above-mentioned criteria, we propose to add three penalties:
1) the value of the center element of a convolution should be large for the second criteria; 2) the values of other element of a convolution (except the center one) should be diverse (i.e., not all of them are similar) for the first criteria; and 3) an additional $L_2$ regularization to avoid overfitting. The three penalties for all convolutions (denoted as $fs$) are derived as follows, respectively.

\vspace{-4mm}
\begin{center}
\begin{align*}
    L_{center} &= -\sum_{f \in fs} \sum_{c \in f.channels} c.center \\
    L_{div} &= -\sum_{f \in fs} \sum_{c \in f.channels} H\big\{\delta(c \setminus c.center)\big\} \\
    L_2 &= \sum_{f \in fs} \sum_{c \in f.channels} \sum_{e \in c} e^2
\end{align*}
\end{center}

\noindent where $f.channels$ denotes the channels of a convolution $f$ and $c.center$ denotes the center element of a channel $c$ of a convolution. Here we first use Softmax function $\delta(z)=\frac{e^{z_i}}{\sum_{z_j \in z} e^{z_j}}$ to normalize the values of elements in a channel $c$ and then apply Shannon index $H(z)=-\sum_{z_i \in z} (z_i \ln z_i$) to model the diversity. The overall penalty is obtained by weighted sum of the three penalties as.

\begin{center}
\vspace{-4mm}
\begin{equation}
    L_{penalty} = \gamma_{1} L_{center} + \gamma_{2} L_{div} + \gamma_{3} L_2
\end{equation}
\end{center}

\subsection{Temporal Layer}

After transforming a path $T$ as a sequence of spatial moving patterns $S$ in the previous layer, DeepIST captures the temporal dependency among $S$ in the temporal layer for travel time estimation of $T$. As mentioned earlier, although it is natural to consider recurrent neural network models (RNN), e.g., LSTM, which have received great success in modeling sequence data in natural language processing~\cite{palangi2016deep} and in image/video processing~\cite{srivastava2015unsupervised}, they typically suffer from vanishing gradient and exploding gradient issues.
Later we experimentally show CNN models perform better than RNN for travel time estimation. More specifically, we propose to apply a one-dimensional CNN (1D-CNN) inspired by~\cite{kim2014convolutional} to capture the local temporal dependency along $S$ instead of trying to capture the temporal dependency for the whole $S$ as an RNN model does. Our idea is based on the intuition behind the First Law of Geography~\cite{tobler1970computer}, ``near things are more related than distant things''. Specifically, 1D-CNN, with multiple convolutional layers, first applies the convolution operation on $S$ along with a one-dimensional sliding window to capture the local temporal patterns among spatial patterns in each window, and use the later convolutional layers further capture higher-level temporal patterns hierarchically. 

\begin{figure}[t]
	\centering
	\includegraphics[scale=0.46]{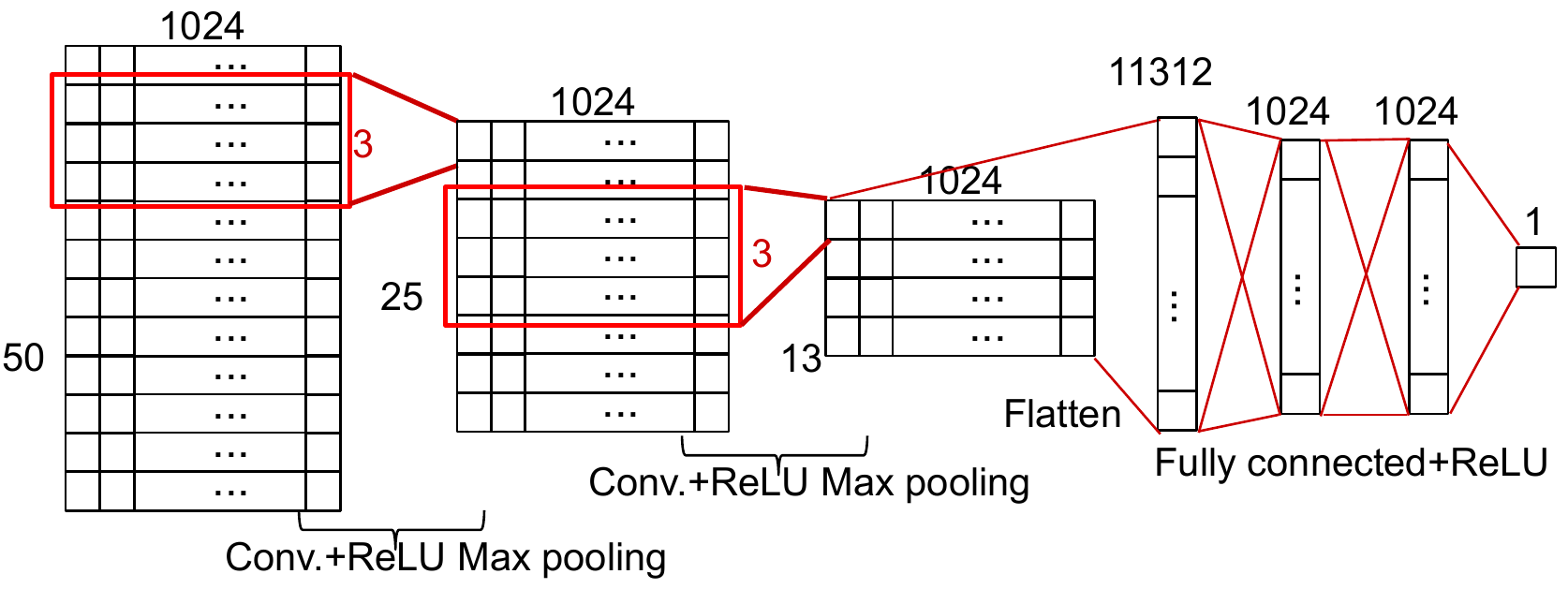}
	\caption{Temporal Layer}
	\label{fig:temporal}
	\vspace{-8mm}
\end{figure}

\noindent{\bf Model Architecture. }As shown in Figure~\ref{fig:temporal}, the proposed 1D-CNN model contains $N$ convolution layers (e.g., $N$=2 in Figure~\ref{fig:temporal}), followed by a flatten layer and $r$ stacked fully-connected layer for the final estimation (e.g., $r$=3). More specifically, for a given $S \in R^{|I| \times \lambda}$, the 1D-CNN model first truncates it by a maximal length $S_{max}$ if $|S|>S_{max}$ (or appends zero vectors if $|S|<S_{max}$) as $S' \in R^{S_{max} \times \lambda}$. $S_{max}$ is determined by data while it can cover more than 99\% spatial pattern sequences of all historical paths, (e.g., $S_{max}$=50 and $\lambda$=1024 in Figure~\ref{fig:temporal}). Then, the 1D-CNN takes $S'$ as input $y^0$ and feeds $y^0$ into $N$ one-dimensional convolutional layers. In each convolutional layer, it applies the convolution operation on the sequence $y^n$ along with a one-dimensional sliding window, and uses padding for location at boundaries of $y^n$ to maintain the dimension of representations. The transformation at each convolutional layer $n$ is derived as follows.

\begin{center}
\vspace{-4mm}
\begin{equation}
    y^n = max\_pooling\big(f(y^{n-1} \ast' W^n_{1D} + b^n_{1D})\big)
\end{equation}
\end{center}

\noindent where $\ast'$ denotes the one dimensional convolutional operation and $f(\cdot)$ is an activation function. Here, we also use the ReLU function. We apply $max\_pooling(\cdot)$ to reduce the dimensionality of the representations after the convolution operation. The number of convolutions in each convolutional layer is denoted as $c_{1D}^n$ and the size of each convolution is denoted as $f_{1D}$, (e.g., $c_{1D}^n$=1024 and 1024 for $n$=1 and 2, respectively and $f_{1D}$=3 shown in Figure~\ref{fig:temporal}).

After $N$ convolution layers, we use a flatten layer to transform the output $y_N$ to a feature vector $\overline{t}$, which is taken as the input $t^0$ for the following $r$ stacked fully-connected layers to estimate the final result $\hat{t}$, i.e., the travel time estimation of the given trajectory $T$, derived as $t^i = f\big(W^i t^{i-1} + b^i\big)$, where $W^i$ and $b^i$ are the weights and biases of the $i^{th}$ fully-connected layer. As shown in Figure~\ref{fig:temporal}, the dimension of the $r=3$ stacked fully-connected layers in our design are 1024, 1024 and 1, respectively, and the final estimation $\hat{t}=t^r$. Note that we also adopt a dropout mechanism in the fully-connected layers.

\subsection{Multi-task Learning}

We introduce a multi-task learning in the DeepIST framework. During the training phase, DeepIST aims to not only estimate the travel time for entire path but also for each sub-path simultaneously. During the test phase, DeepIST skips the sub-path estimation part and returns the estimated the travel time for the entire path only. More specifically, for a spatial pattern sequence $S=\{s_1, ..., s_n\}$ transformed from $T$, each $s_i$ corresponds to the spatial features of sub-path $\tau_i$. We simply use two stacked fully-connected layers with size $\lambda$ and 1 to map each $s_i$ to a scalar to estimate the travel time of the sub-path. The travel time (ground truth) of each sub-path is calculated based on map matching results. More specifically, after matching a trajectory on a road network, each GPS sample point is re-located on a road segment. We simply assume that the moving speed between two consecutive sample points (there may be multiple road segments between them) is constant. Based on this, 
the travel time of any sub-path along the path is derived.


\subsection{Loss Function}

We finally present the loss function of DeepIST, which is trained end-to-end. During the training phase, we use the mean absolute percentage error (MAPE) as our objective functions for travel time estimation of both entire path and each sub-path, which is a relative error to lead DeepIST to estimate accurate results for both the short paths and the long paths. Given a training dataset $D$, for the entire path estimation, the loss function is derived as follows.

\begin{center}
\vspace{-4mm}
\begin{equation}
    L_{path}(\theta) = \sum_{T \in D} \frac{|\hat{t}_{T} - time(T)| }{time(T)}
\end{equation}
\end{center}

\noindent where $\theta$ denotes the trainable parameters in DeepIST, $\hat{t}_{T}$ denotes the final output as the estimated travel time by DeepIST and $time(T)$ is the ground truth for travel time of $T$, which is the time interval between the timestamps of the last and first sample points in $T$. For the sub-path estimation, the loss function is derived as follows.

\begin{center}
\vspace{-4mm}
\begin{equation}
    L_{sub}(\theta) = \sum_{T \in D} \sum_{\tau \in T} \frac{ |\hat{t}_{\tau} - time(\tau)|}{time(\tau)}
\end{equation}
\end{center}

\noindent where $\tau$ denotes a sub-path of $T$, $\hat{t}_{\tau}$ denotes the estimated travel time of $\tau$ and $time(\tau)$ is the ground truth for $\tau$. Our model is trained to minimize the weighted combination of two loss terms corresponding with penalties in PathCNN.

\begin{center}
\vspace{-4mm}
\begin{equation}
    L = \beta L_{path} + (1-\beta)L_{sub} + \gamma_{1} L_{center} + \gamma_{2} L_{div} + \gamma_{3} L_2
\end{equation}
\end{center}

\noindent where $\beta$ is for weighting $L_{path}$ and $L_{sub}$.
\section{Experiments}

In this section, we conduct extensive experiments using two large-scale real-world trajectory datasets to evaluate the performance of DeepIST against several state-of-the-art methods for travel time estimation. We also perform sensitivity tests on parameters of DeepIST, examine several issues in DeepIST.

\subsection{Datasets}
The following describes the trajectory datasets used in the evaluation. For each dataset, we first select trajectories with travel time within 1 to 60 minutes and match the trajectories to the road network by existing map matching techniques~\cite{newson2009hidden, barefoot} to obtain the corresponding sequence of road segments (i.e., paths). Then we filter mismatched paths which have much longer (+10\%) or shorter (-10\%) moving distances comparing with their raw trajectories. Some statistics of the extracted paths are summarized in Table~\ref{table:datasets}.

\begin{table}[]
\centering
\caption{Statistics of datasets}
\label{table:datasets}
\begin{tabular}{|c|c|c|}
\hline
Dataset                & Porto      & Chengdu    \\ \hline \hline
\# of paths     & 700,626  & 2,016,782 \\ \hline
moving distance mean & 6.373 km     & 5.398 km  \\ \hline
travel time mean       & 694.259 sec & 777.879 sec \\ \hline
\end{tabular}
\vspace{-5mm}
\end{table}

\noindent {\bf Porto} taxi data, made available for the Taxi Service Trajectory Prediction Challenge@ ECML/PKDD 2015~\cite{porto}, contains taxi trajectories of 442 taxis running from January 2013 to June 2014 in Porto, Portugal. We generate paths by the above-mentioned pre-processing method to yield 0.7 million paths.

\noindent {\bf Chengdu} taxi data, made available for the Taxi travel time estimation challenge@ dcjingsai 2016~\cite{chengdu}, contains 1.4 billion GPS sample points of 14,864 taxis running in August 2014 in Chengdu, China. We segment the sample points of each taxi into trajectories based on a 60-second time gap. Finally, we generate paths by the above-mentioned pre-processing method to yield 2.02 million paths.

\subsection{Baseline Methods for Comparison}
The travel time estimation methods we evaluate for comparison include 
five state-of-the-arts (road-based, path-based and three learning-based methods) and one variant of our proposed method.

\begin{table*}[]
\centering
\caption{Performance Evaluation}
\label{table:evaluation}
\begin{tabular}{|c|c|c|c|c|c|c|}
\hline
Dataset          & \multicolumn{3}{c|}{Porto}                                                    & \multicolumn{3}{c|}{Chengdu}                                                  \\ \hline \hline
Metric           & RMSE (sec)               & MAE (sec)                & MAPE (\%)               & RMSE (sec)               & MAE (sec)                & MAPE (\%)               \\ \hline \hline
spd-LSTM         & 281.94                   & 127.61                   & 18.58                   & 406.94                   & 185.44                   & 23.96                   \\ \hline
TEMP             & 254.49                   & 115.59                    & 16.71                   & 341.57                   & 147.09                   & 18.84                   \\ \hline
DeepTravel       & 228.32                    & 98.59                    & 14.01                    & 228.56                   & 125.78                    & 16.25                   \\ \hline
WDR          & *159.35                   & *70.67                   & *10.25                   & 238.85                  & 105.71                   & 13.74                  \\ \hline
DeepTTE          & 180.41                  & 75.47                  & 10.92                   & *212.74                  & *95.52                   & *12.53                  \\ \hline
DeepIST$_{LSTM}$ & 112.64                    & 95.29                    & 8.67                    & 185.63                    & 78.02                    & 10.13                    \\ \hline
DeepIST          & \textbf{98.22 (-38.36\%)} & \textbf{53.45 (-24.37\%)} & \textbf{7.78 (-24.06\%)} & \textbf{167.81 (-21.12\%)} & \textbf{71.02 (-25.64\%)} & \textbf{9.45 (-24.58\%)} \\ \hline
\end{tabular}
\end{table*}


\noindent {\bf spd-LSTM~\cite{ma2015long}}, a road-based method, uses LSTM model to predict the speed of each road segment by using historical travel speeds with hourly time slots. The travel time of a query path is estimated by summing up the travel time of road segments in it. 

\noindent {\bf TEMP~\cite{wang2016simple}}, a path-based method, estimates the travel time of a given path based on the nearby historical trajectories, which have close source and destination with the query path. About 9\% of the query paths can not be estimated in the original TEMP method due to the lack of nearby trajectories. For those paths, we enlarge the neighborhood in TEMP method until finding enough number of nearby trajectories (i.e., 10 trajectories in our experiments).

\noindent {\bf DeepTravel~\cite{zhang2018deeptravel}}, a learning-based method, transforms a query path into a sequence of geographic cells, each of which is represented by several manually-craft features. LSTM model is used to learn and estimate the travel time of the query path.

\noindent {\bf WDR}~\cite{wang2018learning}, a learning-based method, transforms a query path as a number of manually-craft features. An ensemble regression model, consisting of a wide, a deep fully-connected network and an LSTM model, is used to estimate the travel time of the query path.

\noindent {\bf DeepTTE~\cite{wang2018will}}, a learning-based method, transforms a query path as a sequence of windows containing consecutive re-sampled points along the path (with equal distance gaps) and extracts $k$ features for each window. LSTM model is used to learn and estimate the travel time of the query path.

\noindent {\bf DeepIST$_{LSTM}$}, a variant of DeepIST, uses LSTM model instead of 1D-CNN model in the temporal layer of the original DeepIST.

\subsection{Experimental Setup}

In each experiment, for a dataset, we randomly split it into three folds, 80\%, 10\% and 10\% as the training set, the validation set, and the test set, respectively. We use the training set to train models while using the validation set to select the best models, and evaluate the performance using the test set. We repeat each experiment for 5 times and report the mean of the different runs. We use {\em mean absolute error (MAE)}, {\em mean absolute percentage error (MAPE)} and {\em root-mean-squared error (RMSE)} between the predicted results and the ground truths (in seconds) as the performance metrics. 

Regarding the default parameter settings, for DeepIST, the details for network architecture are shown in Figure~\ref{fig:spatial} and Figure~\ref{fig:temporal}. The size of convolutions in both PathCNN and 1D-CNN is set to $3\times3$. The window size $w$ of the sliding window is set to 0.5 km, and the sliding step $s$ is set to 0.4 km to generate the overlapped windows. The $S_{max}$ is determined by data while it can cover more than 99\% spatial pattern sequences of paths, (e.g., $S_{max}$=53 for the Porto dataset). The image size $k \times k$ for each window is set to $100 \times 100$, and each image covers the same size of geographic area of $0.5 \times 0.5 (km^2)$ (i.e., $r_{lng}$=0.0058699 and $r_{lat}$=0.0044966), which reflect the granularity of an image. The $\beta$ which balances the error between estimation of the entire path and sub-paths is set to 0.6, and the $\gamma_1$, $\gamma_2$ and $\gamma_3$ for weighting the regularization is set to 0.1, 0.1 and 0.01, respectively. For DeepIST$_{LSTM}$, we set the hidden unit as 1024 for the singly-layer LSTM model in temporal layer, and use the same parameter settings used in DeepIST for other parameters. For DeepTravel, WDR and DeepTTE, we generate the required configurations and meta-data for each dataset and tune the best parameter settings. The initial learning rate is set to 0.0001 and we use Adam~\cite{kingma2014adam} for optimization. To obtain converged results, the number of iterations for model training varies for individual models and different datasets.

Our model is implemented in python with Tensorflow. We train and evaluate all models on a server with NVIDIA GTX 1080 GPU, and one Intel Core i5-8400 CPU on the Ubuntu 18.04.

\vspace{-5mm}
\subsection{Evaluation of Models}

The performance obtained by all evaluated methods is summarized in Table~\ref{table:evaluation}. DeepIST outperforms all the compared methods. As shown, the improvement ratio of MAE (compared with the best of these existing models, marked by '*') are 24.37\% and 25.64\% in Porto and Chengdu datasets, respectively. We have the following observations from the comparison.

\noindent {\bf Learning based models outperform road- and path-based methods.} Comparing with non-learning-based methods, spd-LSTM and TEMP, the learning-based models (i.e., DeepTraval, WDR, DeepTTE, \\ DeepIST$_{LSTM}$ and DeepIST) achieve better performance because they can learn discriminative patterns from the whole paths for the estimation while spd-LSTM may miss the correlations and intersections between road segments and TEMP fails to find nearby paths of the query paths or its assumption fails.

\noindent {\bf PathCNN is effective to extract spatial moving patterns.} Among DeepTravel, WDR, DeepTTE and DeepIST$_{LSTM}$, all of which exploit LSTM to capture temporal patterns among extracted moving patterns while exploring different approaches to extract spatial moving patterns, DeepIST$_{LSTM}$ achieves the best performance, indicating that it is more effective than others in extracting spatial moving patterns by exploring the novel ideas in PathCNN. 

\noindent {\bf Exploring 1D-CNN is effective to capture temporal patterns.} Comparing with DeepIST$_{LSTM}$, 
the performance improvement of DeepIST is clear and impressive. It shows that the proposed 1D-CNN is more effective than LSTM in capturing local temporal patterns for travel time estimation.

\begin{figure}
\centering
  \subfigure[Porto]{
    \centering
    \includegraphics[width=0.47\linewidth]{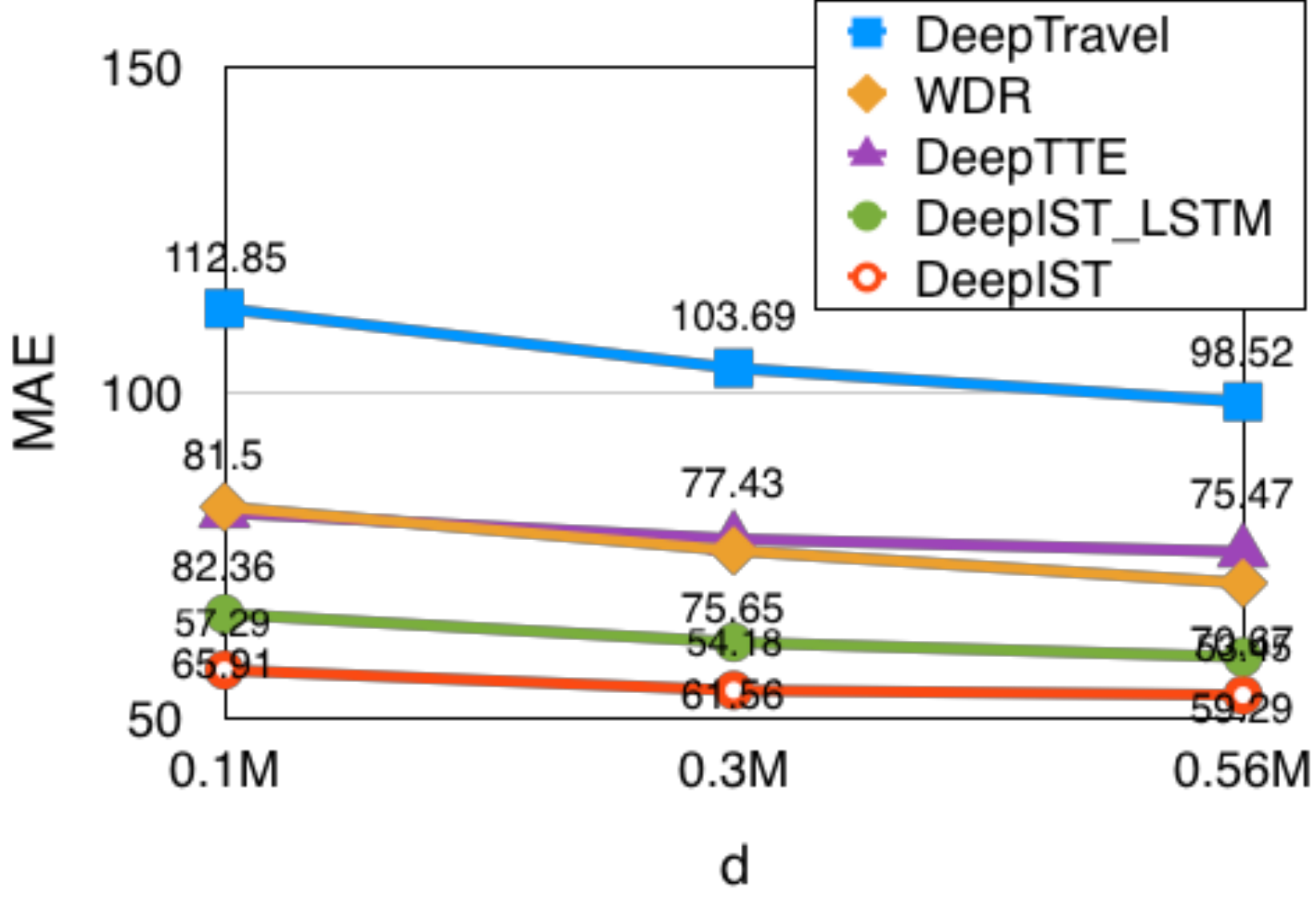}
  }
  \subfigure[Chengdu]{
    \centering
    \includegraphics[width=0.47\linewidth]{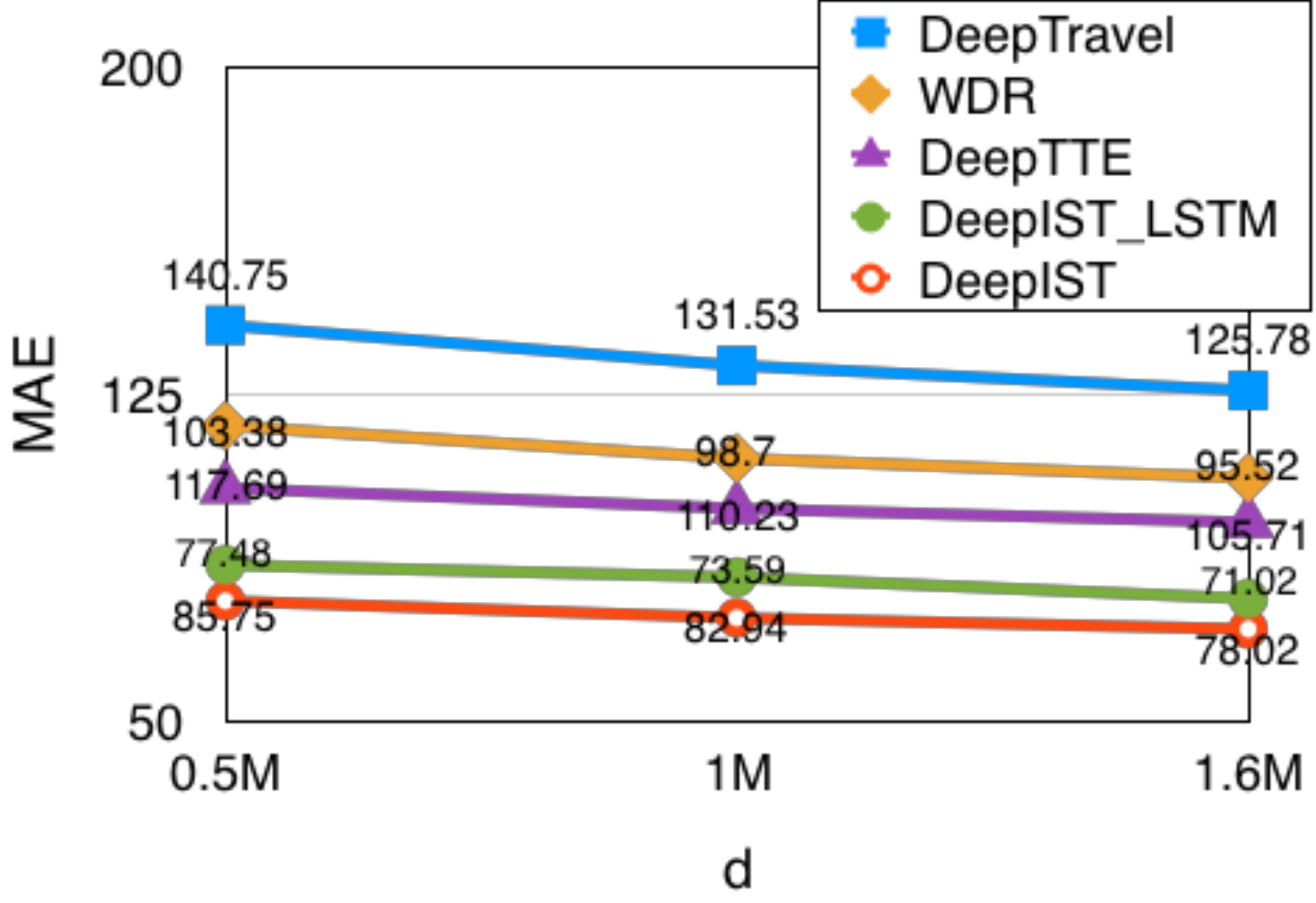}
  }
  	\vspace{-3mm}
  \caption{Training Set Size}
  	\vspace{-5mm}
\label{fig:d}
\end{figure}

Here, we also study the impact of training set size for learning-based models. In the following, we only show MAE for evaluation because RMSE and MAPE show similar trends with MAE. For the training set size, we increase the training set size $d$ from 0.1M to 0.56M for Porto dataset and 0.5M to 1.6M for Chengdu dataset. As shown in Figure~\ref{fig:d}(a) and Figure~\ref{fig:d}(b) for both datasets, as the training size $d$ increases, the performance of all learning-based models continue to improve and converge when $d$ is greater than $0.3M$ in Porto and $1M$ in Chengdu. It also shows DeepIST soundly outperforms all other models in both datasets under various $d$.

\subsection{Parameter Sensitivity in DeepIST}

\begin{figure}
\centering
  \subfigure[No. of Layers of PathCNN]{
    \centering
    \includegraphics[width=0.47\linewidth]{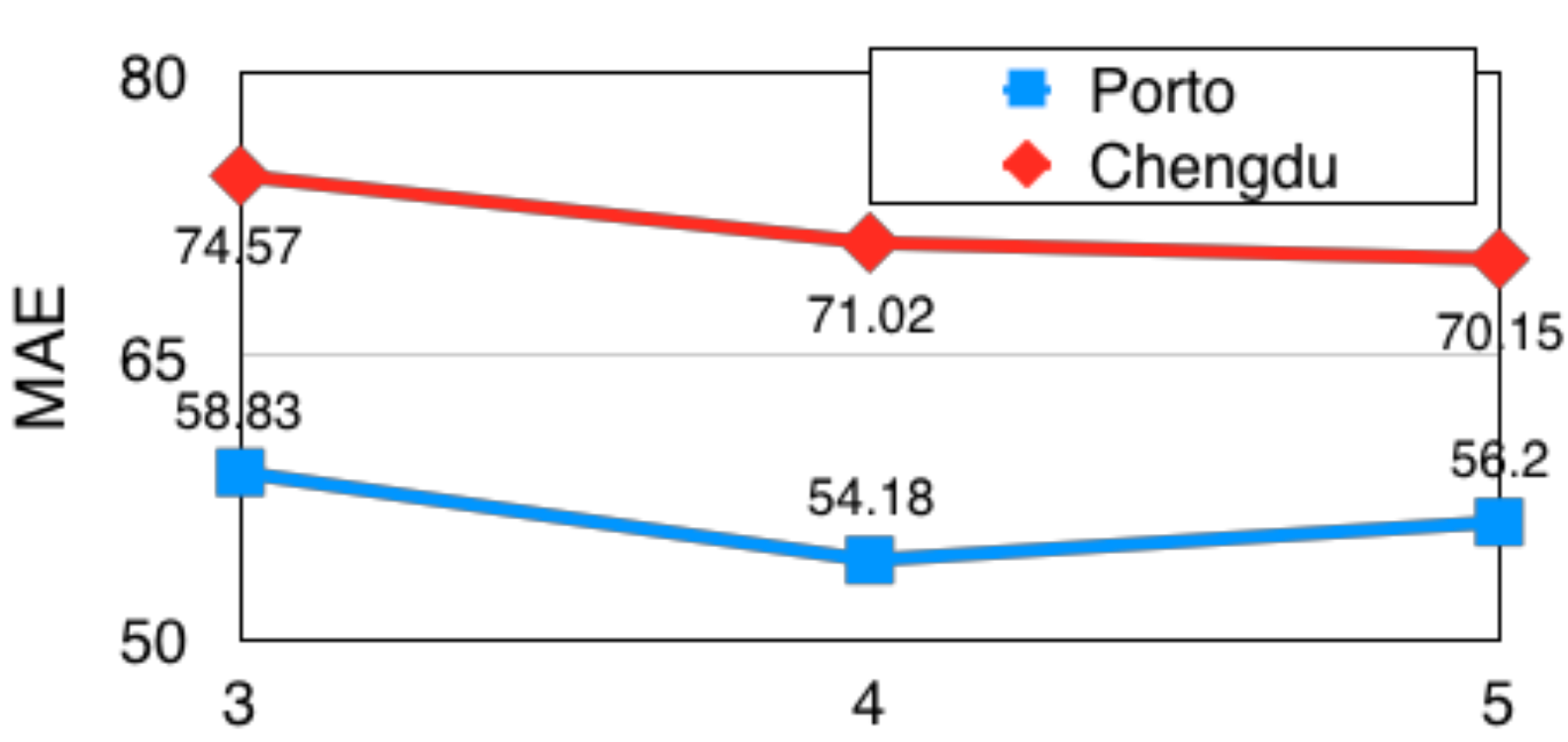}
  }
  \subfigure[No. of Convolutions of PathCNN]{
    \centering
    \includegraphics[width=0.47\linewidth]{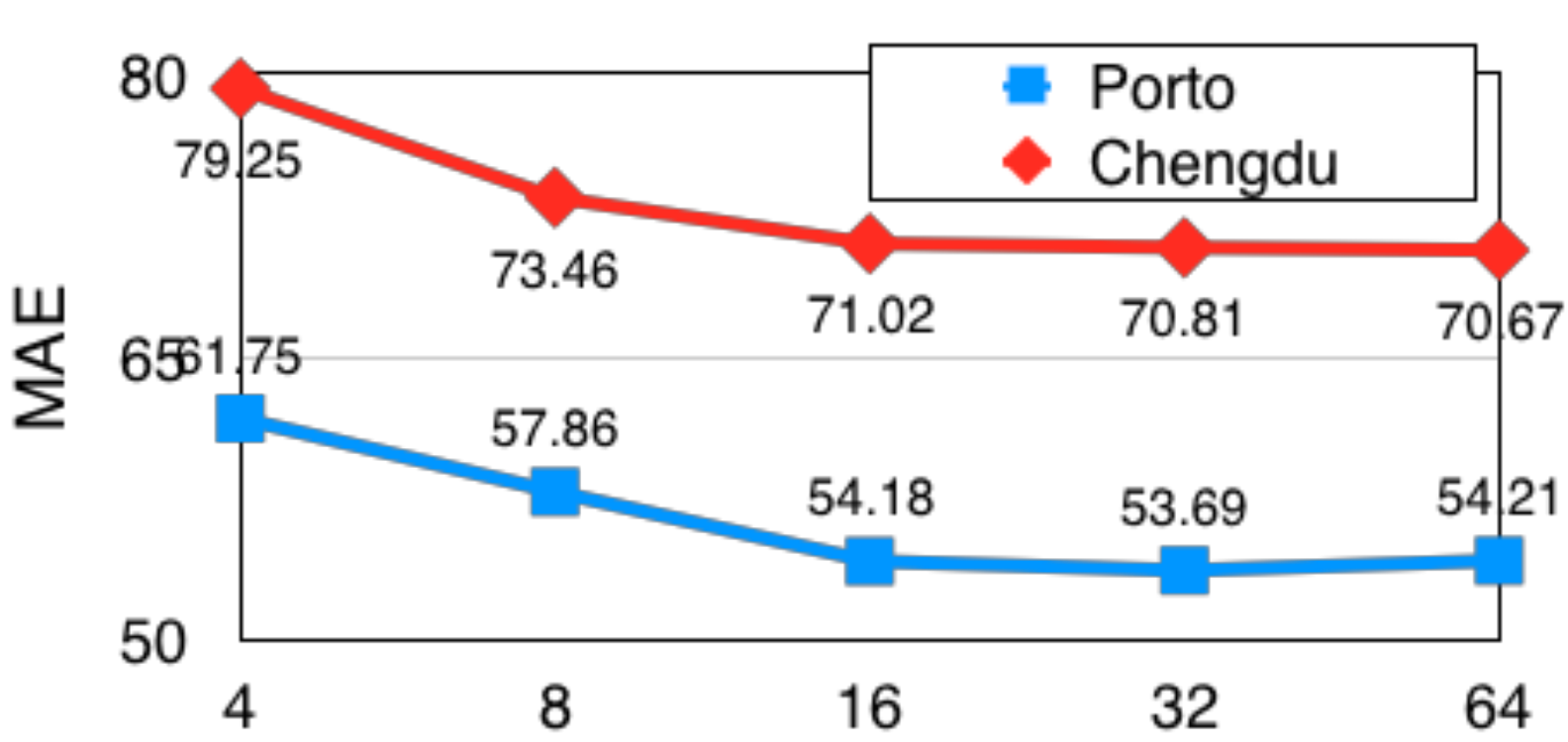}
  }
  \\
  \subfigure[No. of Layers of 1D-CNN]{
    \centering
    \includegraphics[width=0.47\linewidth]{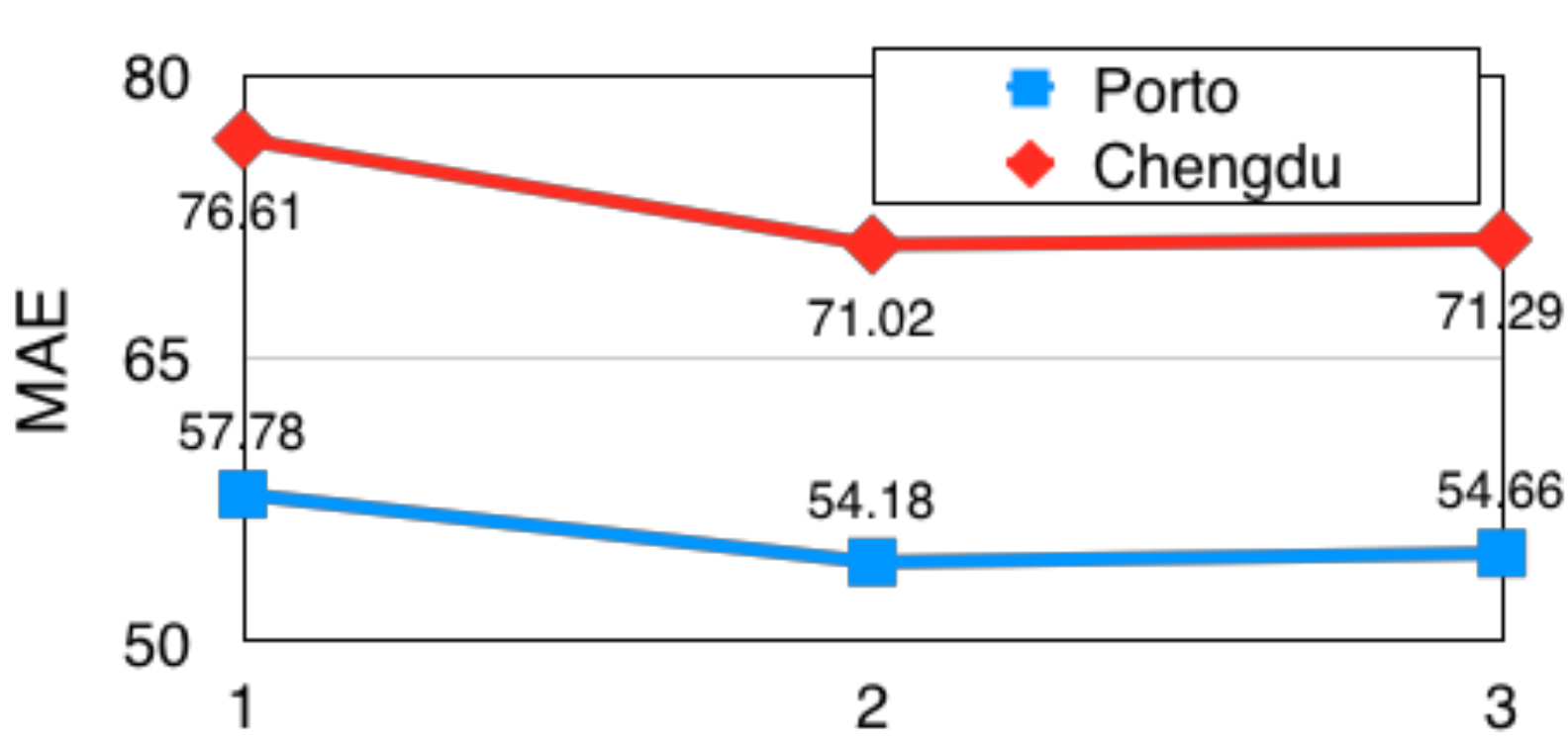}
  }
  \subfigure[No. of Convolutions of 1D-CNN]{
    \centering
    \includegraphics[width=0.47\linewidth]{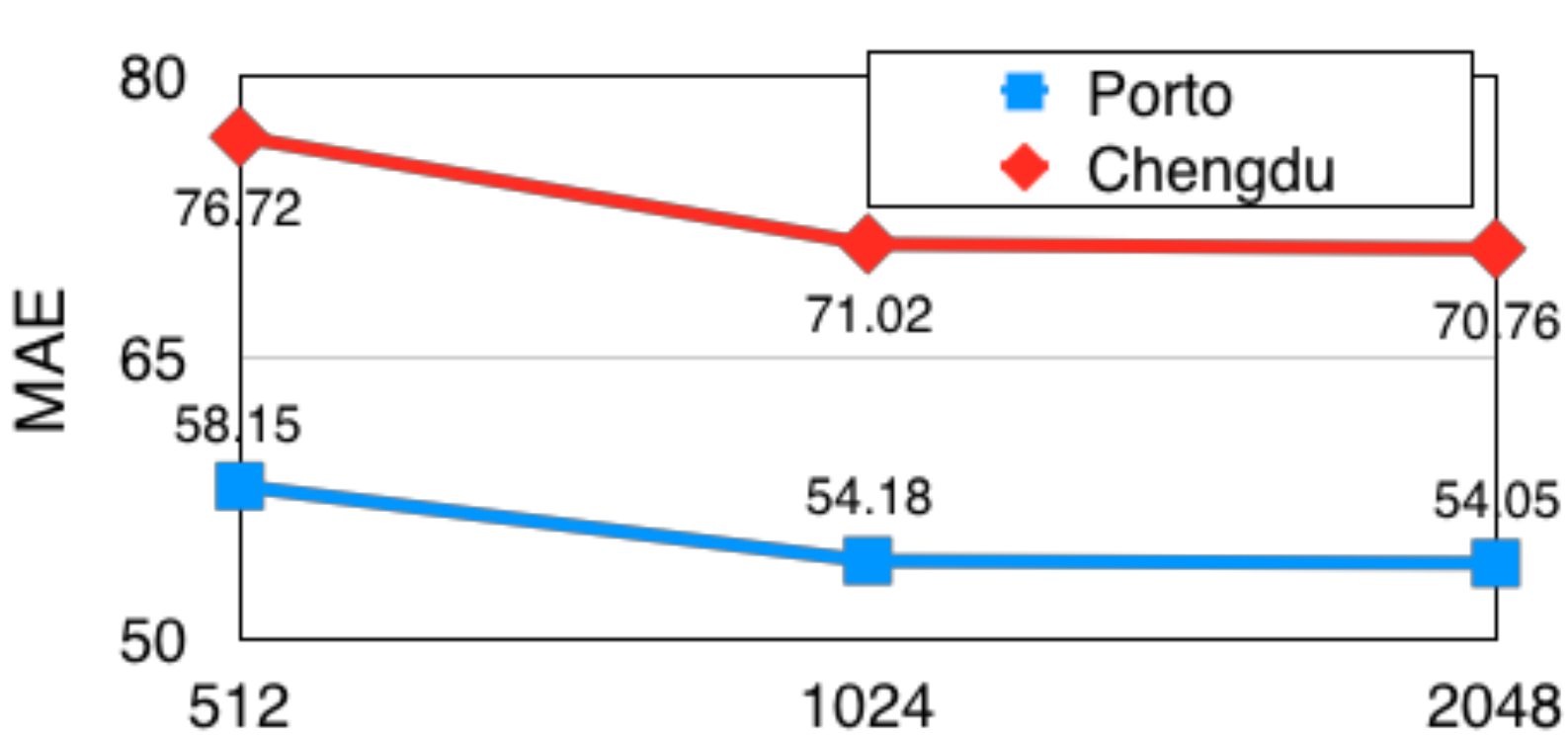}
  }
  \\
  \subfigure[Window Size $w$]{
    \centering
    \includegraphics[width=0.47\linewidth]{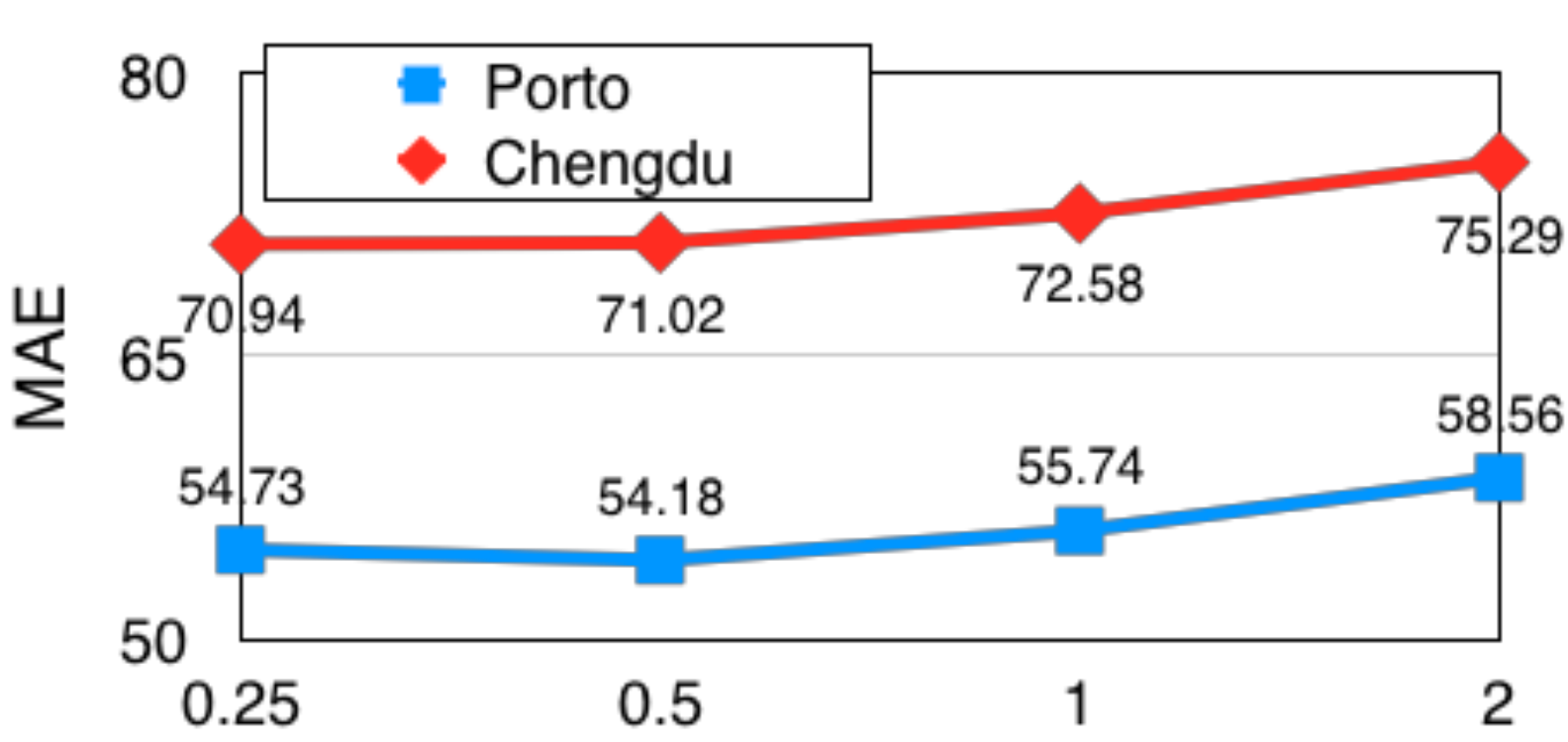}
  }
  \subfigure[Image resolution $k \times k$]{
    \centering
    \includegraphics[width=0.47\linewidth]{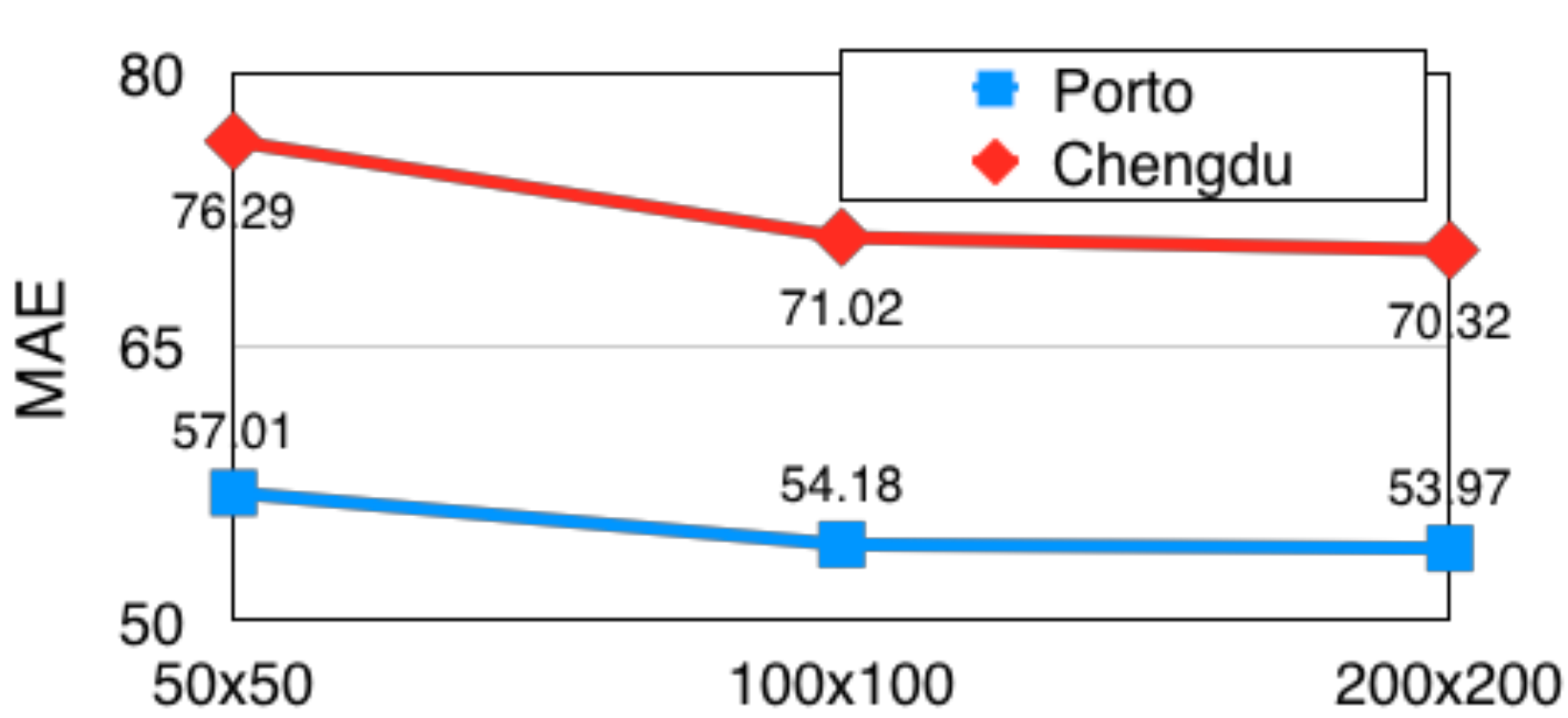}
  }
  	\vspace{-3mm}
  \caption{Parameter Sensitivity of DeepIST}
  	\vspace{-4mm}
\label{fig:parameter}
\end{figure}

Here we examine the impact of parameter settings in DeepIST on its performance. To test the parameter sensitivity of DeepIST, we vary the values of important parameters, the model architecture including the number of layers and the number of convolutions in both PathCNN and 1D-CNN, the window size $w$ and the image size $k \times k$ (which reflects the resolution of images) to observe the changes in MAE, as shown in Figure~\ref{fig:parameter}.

\noindent {\bf Model Architecture.} First of all, we study the impact of the model architecture. Generally speaking, large numbers of layers and convolutions describe a deep and wide network, which is capable to fit the training data, but may need large size of training data. On the other hand, small numbers of layers and convolutions desribing a shallow and narrow network which is not capable to generalize the training set for the estimation. As shown in Figure~\ref{fig:parameter}(a)-(d), for both datasets, the best performance is achieved when number of layers in PathCNN and 1D-CNN are set to 4 and 2, respectively, and the performance does not change much when the number of convolutions of PathCNN and 1D-CNN are larger than 16 and 1024.

\noindent {\bf Window size $w$.} Figure~\ref{fig:parameter}(e) shows that setting the window size of the sliding window smaller than 0.5 km in both Porto and Chengdu datasets achieves the best performance. A small window size $w$ forces the model to focus on extracting local spatial moving patterns from shorter sub-paths while a large window size may cover irrelevant points and may lead to noises and cause overfitting.

\noindent {\bf Image resolution $k \times k$.} Figure~\ref{fig:parameter}(f) shows that setting the resolution of images greater than $100 \times 100$ for both Porto and Chengdu datasets achieves converged performance. Generally speaking, a small $k$, indicating low resolution in images, does not provide sufficient information of paths in images. On the other hand, the maximum size of images is constrained by hardware
and a large $k$ leads to long training time.

Based on these results, in DeepIST, the number of layers in PathCNN and 1D-CNN is set to 4 and 2, respectively. The number of convolutions in PathCNN and 1D-CNN is set to 16 and 1024, respectivly. The window size $w$ is set to 0.5 km, and the resolution $k \times k$ of generated images is set to $100 \times 100$. There are several additional parameters in DeepIST that are empirically decided, including convolution size, sliding step, $\beta$ and weights for regularization in PathCNN, in the process of tuning the above parameters.
Due to the lack of space, we skip the details in tuning those parameters.

\subsection{Study of Unique Issues in DeepIST}
\label{sec:issue}

In this section, we examine the following unique issues arising in the design of DeepIST: i) the effect of overlapped windows, ii) the effect of plotting traffic condition, road network, and traffic signals in images, iii) the effect of alternative designs of max+avg layer in PathCNN and iv) the effect of multi-tasking and regularization in PathCNN. We perform experiments to compare alternative choices in these issues and justify our decisions. 

\begin{figure}
\centering
  \subfigure[Plotted on Images]{
    \centering
    \includegraphics[width=0.47\linewidth]{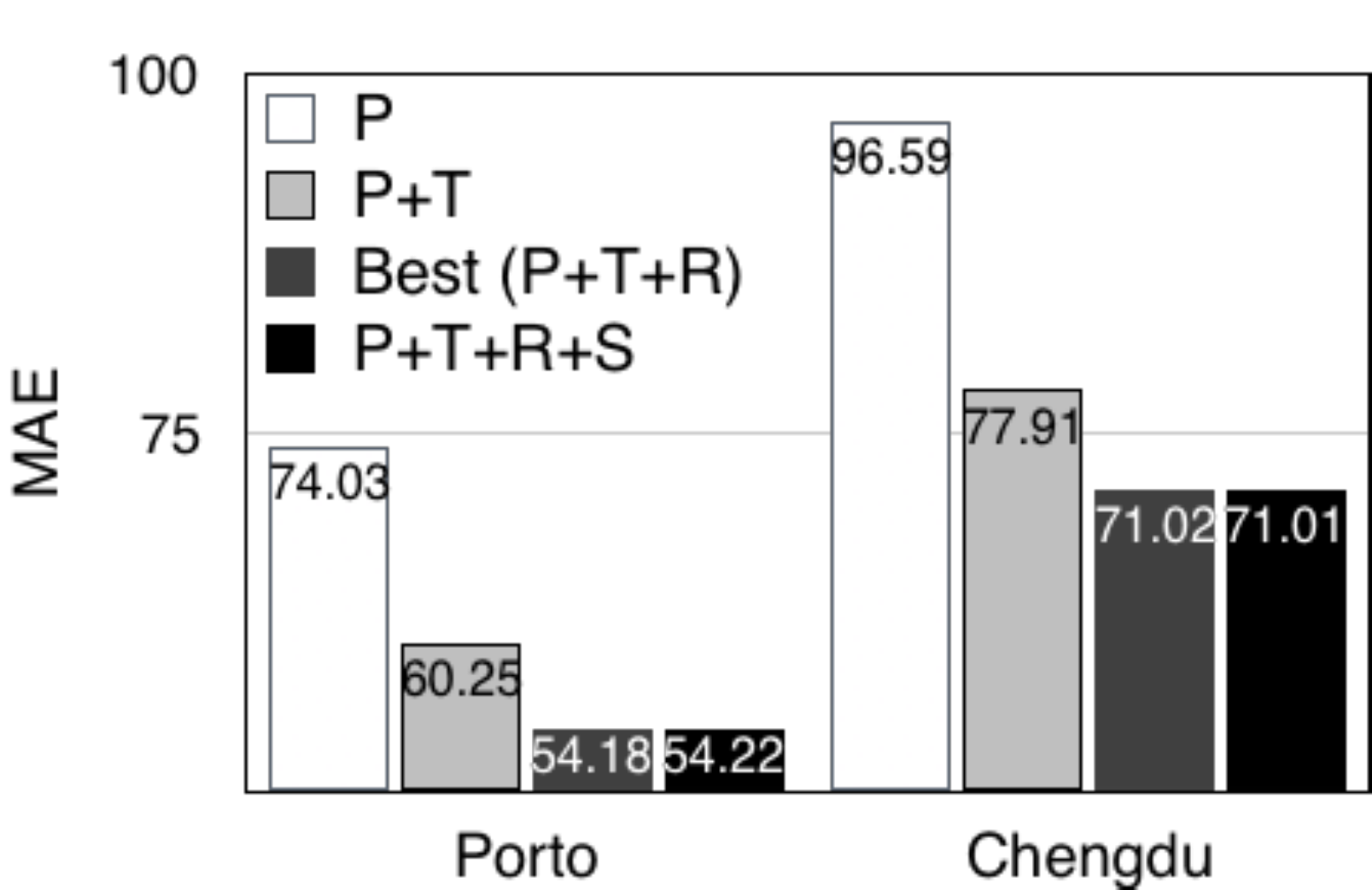}
  }
  \subfigure[Overlaps of Windows]{
    \centering
    \includegraphics[width=0.47\linewidth]{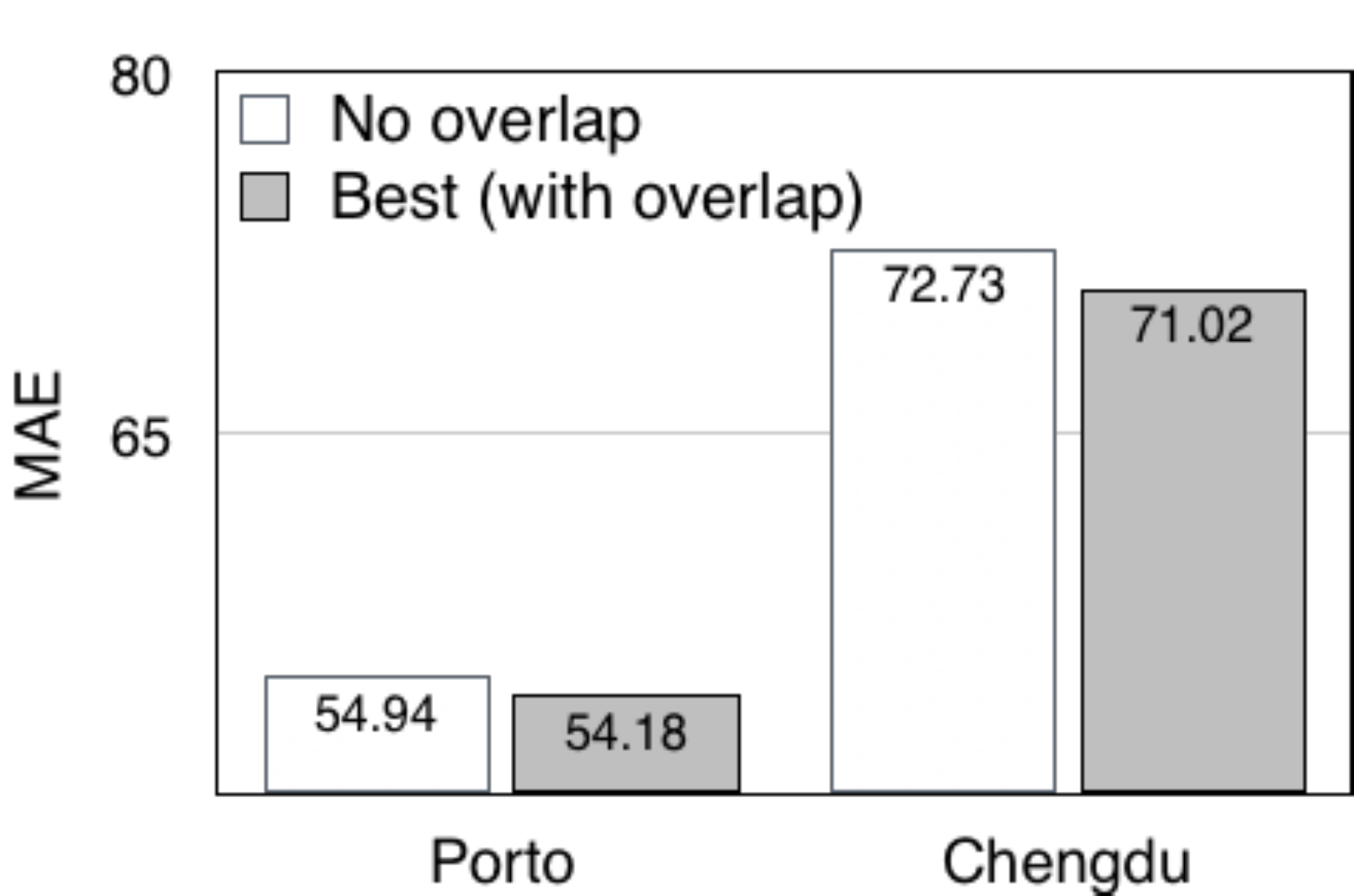}
  }
  \\
  \vspace{-3mm}
  \subfigure[Layer Design of PathCNN]{
    \centering
    \includegraphics[width=0.47\linewidth]{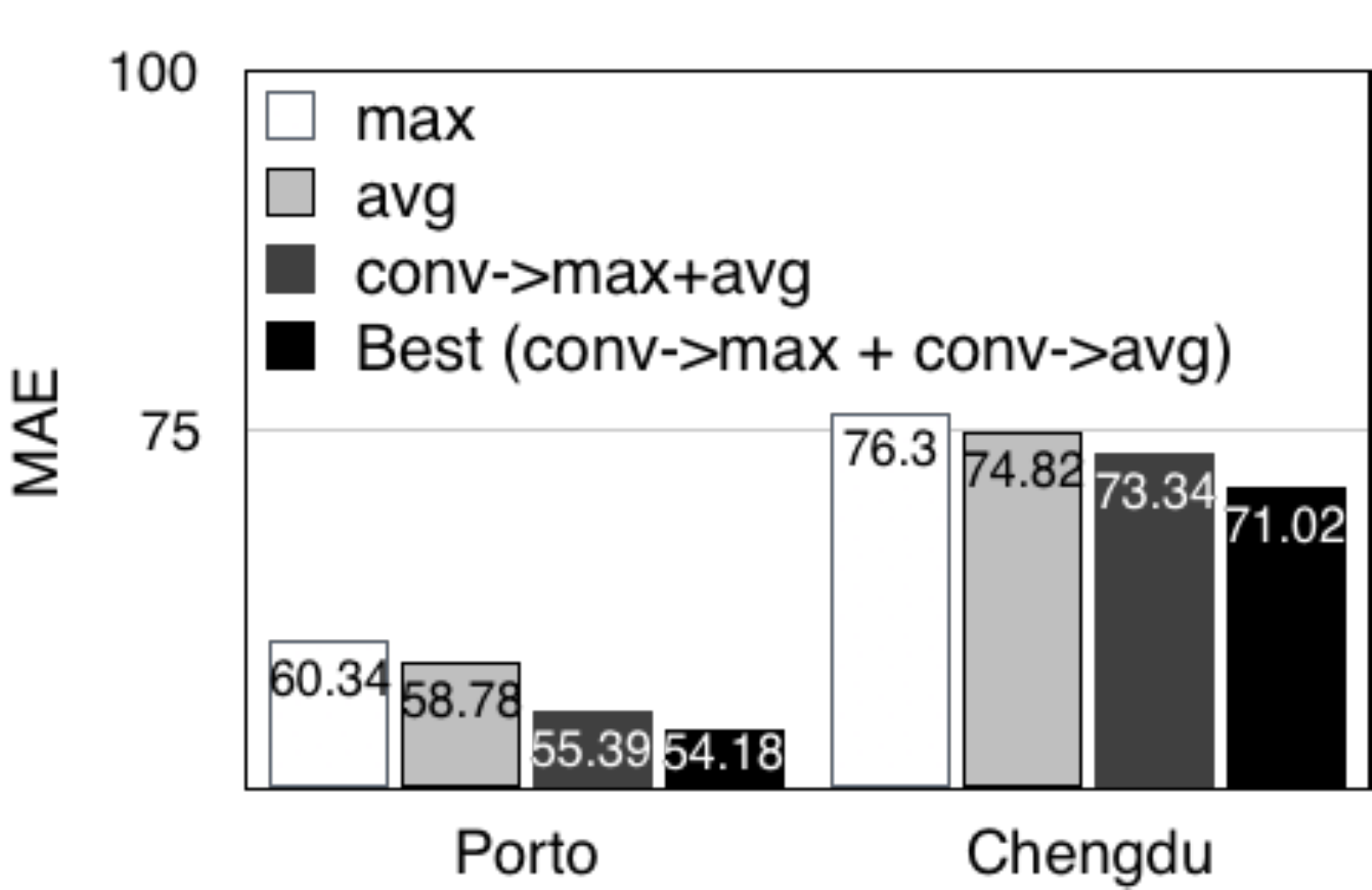}
  }
  \subfigure[Issues of PathCNN]{
    \centering
    \includegraphics[width=0.47\linewidth]{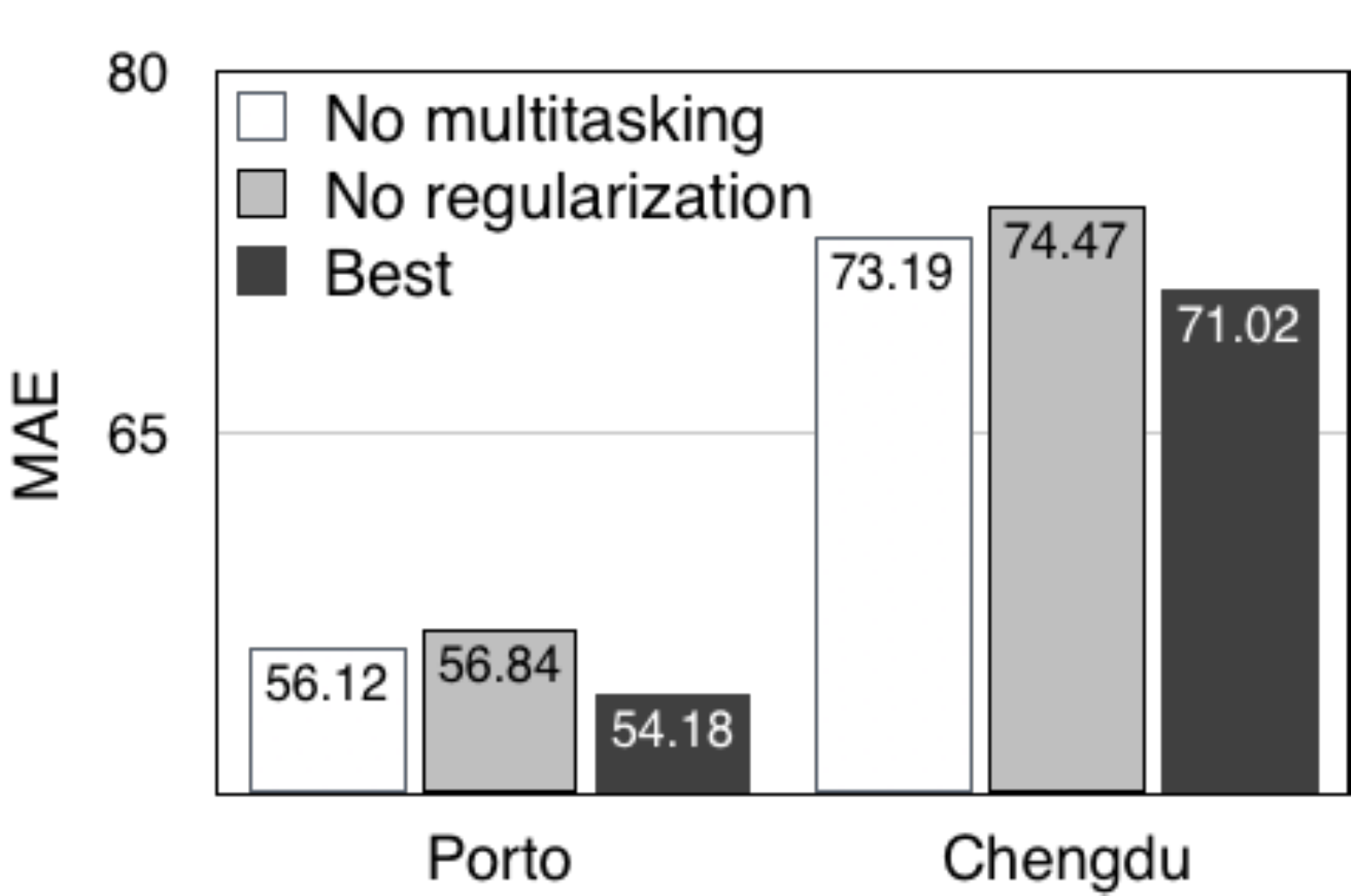}
  }
  	\vspace{-3mm}
  \caption{Comparison of approaches to issues in DeepIST}
  	\vspace{-4mm}
\label{fig:issue}
\end{figure}

Regarding the issue of plotting additional information in the ``generalized'' images, we compare four different approaches: 1) \texttt{P} plots only sub-paths of query paths, 2) \texttt{P+T} further plots the corresponding traffic condition, 3) \texttt{Best (P+T+R)} further plots the corresponding road network and 4) \texttt{P+T+R+S} further plots the corresponding traffic signals.
Figure~\ref{fig:issue}(a) shows that while \texttt{P} already achieves competitive performance with existing works, \texttt{P+T} significantly improves the performance and \texttt{Best(P+T+R)} further improves the performance. This observation suggests that the information of traffic condition and road network are both very useful for the travel time estimation. One the other hand, \texttt{P+T+R+S} does not improve much or even worse than \texttt{Best(P+T+R)}, which may be due to that the incomplete of traffic signal data from the data source, OpenStreetMap (e.g., there are only 2,398 stop signs and 2,447 traffic lights in Porto in OpenStreetMap).

Regarding the issue of overlap in adjacent windows, we compare two settings: 1) the \texttt{No overlap} setting generates non-overlapped windows by setting the same value to the window size and the sliding step, $w$=$s$=0.5km, and 2) the \texttt{Best} generates overlapped windows by setting $w$=0.5km and $s$=0.1km. Figure~\ref{fig:issue}(b) shows that \texttt{Best} outperforms \texttt{No overlap} in both datasets, which suggests that missing moving patterns between adjacent windows as \texttt{No overlap} does is harmful for the travel time estimation.

Regarding the design of max+avg layer of PathCNN, we compare four alternatives: 1) the \texttt{max} adopts a convolution layer followed by a max pooling layer, 2) the \texttt{avg} adopts a convolution layer followed by an average pooling layer, 3) the \texttt{conv->max+avg} adopts one convolution layer followed by both of the max pooling and average pooling layers and 4) \texttt{Best} which is the proposed max+avg layer in PathCNN. Figure~\ref{fig:issue}(c) shows that adopting both max pooling and average pooling layers outperforms adopting only single max pooling layer or average pooling layer. Moreover, \texttt{Best}, which adopts separate two convolution layers, further outperforms \texttt{conv->max+avg}, which adopts a single convolution layer, due to that the single convolution layer would learns aggregated features for the followed two pooling layers.

Finally, we study the effect of multi-tasking learning and regularization in PathCNN for extracting line features. Figure~\ref{fig:issue}(d) shows that \texttt{Best} which adopts both multi-tasking and regularization in PathCNN outperforms \texttt{No multi-tasking}, which does not adopt multi-tasking, and \texttt{No regularization}, which does not adopt regularization in PathCNN. This suggests that the both methods are useful for travel time estimation.


\section{Conclusion}

This study focuses on the travel time estimation for paths. Prior works fail to well capture the spatial moving patterns and temporal patterns among spatial patterns embedded in a path and thus can not estimate the travel time accurately. To fill in this gap, we propose to treat a path as a sequence of sub-paths plotted into images and harness the power of convolutional neural network model (CNN) to model the complex spatial moving patterns and their temporal patterns along the path for travel time estimation. To achieve the goal, we propose a novel three-layer framework, namely DeepIST, which generates images containing sub-paths of paths corresponding with additional information, including traffic conditions, road network and traffic signals, propose a novel two-dimensional CNN, namely PathCNN, to extract spatial patterns for lines in images, and uses a one-dimensional CNN to capture the temporal patterns for travel time estimation. Empirical result shows that DeepIST soundly outperforms all existing models in multiple large-scale real-world datasets. 

\begin{acks}
This work is supported in part by the National Science Foundation under Grant No. IIS-1717084.
\end{acks}

\bibliographystyle{ACM-Reference-Format}
\bibliography{0_main}

\end{document}